\documentclass[sigconf]{acmart}

\usepackage{multirow}
\usepackage{array}
\usepackage{colortbl}
\definecolor{light}{gray}{.85}
\definecolor{title}{gray}{.30}

\AtBeginDocument{%
  }

\copyrightyear{2024}
\acmYear{2024}
\setcopyright{acmlicensed}\acmConference[MM '24]{Proceedings of the 32nd ACM International Conference on Multimedia}{October 28-November 1, 2024}{Melbourne, VIC, Australia}
\acmBooktitle{Proceedings of the 32nd ACM International Conference on Multimedia (MM '24), October 28-November 1, 2024, Melbourne, VIC, Australia}
\acmDOI{10.1145/3664647.3681366}
\acmISBN{979-8-4007-0686-8/24/10}

\begin{document}

\title{MMHead: Towards Fine-grained Multi-modal 3D Facial Animation}

\author{Sijing Wu}
\authornote{Equal contribution.}
\affiliation{%
  \institution{Shanghai Jiao Tong University}
  \city{Shanghai}
  \country{China}
}
\email{wusijing@sjtu.edu.cn}
\orcid{0009-0000-7753-1596}

\author{Yunhao Li}
\authornotemark[1]
\affiliation{%
  \institution{Shanghai Jiao Tong University}
  \city{Shanghai}
  \country{China}}
\email{lyhsjtu@sjtu.edu.cn}

\author{Yichao Yan}
\authornote{Corresponding author.}
\affiliation{%
  \institution{Shanghai Jiao Tong University}
  \city{Shanghai}
  \country{China}}
\email{yanyichao@sjtu.edu.cn}

\author{Huiyu Duan}
\affiliation{%
  \institution{Shanghai Jiao Tong University}
  \city{Shanghai}
  \country{China}}
\email{huiyuduan@sjtu.edu.cn}

\author{Ziwei Liu}
\affiliation{%
  \institution{Nanyang Technological University}
  \country{Singapore}}
\email{ziwei.liu@ntu.edu.sg}

\author{Guangtao Zhai}
\authornotemark[2]
\affiliation{%
  \institution{Shanghai Jiao Tong University}
  \city{Shanghai}
  \country{China}}
\email{zhaiguangtao@sjtu.edu.cn}

\renewcommand{\shortauthors}{Sijing Wu et al.}

\begin{abstract}
3D facial animation has attracted considerable attention due to its extensive applications in the multimedia field. Audio-driven 3D facial animation has been widely explored with promising results. However, multi-modal 3D facial animation, especially text-guided 3D facial animation is rarely explored due to the lack of multi-modal 3D facial animation dataset. To fill this gap, we first construct a large-scale multi-modal 3D facial animation dataset, \textbf{MMHead}, which consists of 49 hours of 3D facial motion sequences, speech audios, and rich hierarchical text annotations. Each text annotation contains abstract action and emotion descriptions, fine-grained facial and head movements (\textit{i.e.}, expression and head pose) descriptions, and three possible scenarios that may cause such emotion. Concretely, we integrate five public 2D portrait video datasets, and propose an automatic pipeline to 1) reconstruct 3D facial motion sequences from monocular videos; and 2) obtain hierarchical text annotations with the help of AU detection and ChatGPT. Based on the MMHead dataset, we establish benchmarks for two new tasks: text-induced 3D talking head animation and text-to-3D facial motion generation. Moreover, a simple but efficient VQ-VAE-based method named \textbf{MM2Face} is proposed to unify the multi-modal information and generate diverse and plausible 3D facial motions, which achieves competitive results on both benchmarks. Extensive experiments and comprehensive analysis demonstrate the significant potential of our dataset and benchmarks in promoting the development of multi-modal 3D facial animation. The dataset will be released at: \url{https://wsj-sjtu.github.io/MMHead/}.
\end{abstract}


\begin{teaserfigure}
  \includegraphics[width=\textwidth]{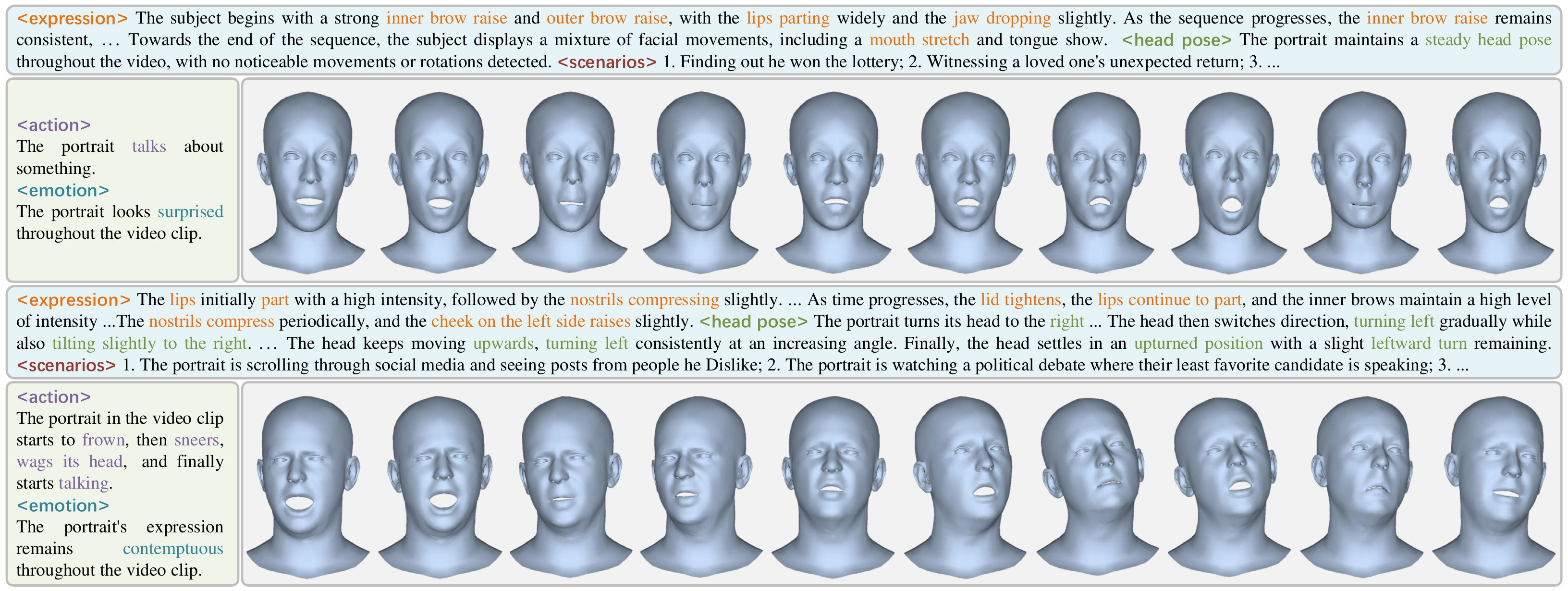}
  \caption{We present MMHead, the first multi-modal 3D facial animation dataset with hierarchical text annotations including abstract descriptions for overall actions and emotions, and fine-grained descriptions for expressions, head poses, as well as possible scenarios that may cause such emotions.}
  \label{fig:teaser}
\end{teaserfigure}


\maketitle

\section{Introduction}

\begin{table*}
\centering
\caption{\textbf{Comparison of relevant datasets.} The abbreviations "Act.", "Emo.", "Exp.", and "Scn." stand for abstract action and emotion annotations, fine-grained expression (facial movement) annotation, and scenarios that may cause such emotion, respectively. The abbreviations "Subj.”, "Dur.", "Lang.", "Pepr.", "Env.", and "Tech." refer to subjects, duration, language, representation, and the data acquisition environment and the technology used, respectively. "L" indicates that the dataset has labels rather than text descriptions. "\checkmark\kern-1.2ex\raisebox{0.7ex}{\rotatebox[origin=c]{125}{\textbf{--}}}" indicates that such annotation exists but is not complete. "EN", "CN", and "Mul." represent English, Chinese, and multiple languages, respectively. "Mono." means the 3D facial motion is reconstructed from monocular video, and "Gen." means the 3D facial motion is generated by neural networks.}
\label{tab:tab1}
\resizebox{\textwidth}{!}{
\begin{tabular}{l|cccc|ccccc|cc|ccc|cc}
\toprule
\multirow{3}{*}{Dataset}   &\multicolumn{4}{c|}{\textbf{Modality}}   &\multicolumn{5}{c|}{\textbf{Annotation}}   &\multicolumn{2}{c|}{\textbf{Scale}}   &\multicolumn{3}{c|}{\textbf{Property}}   &\multicolumn{2}{c}{\textbf{Acquisition}} \\
\cline{2-17}
&\multirow{2}{*}{Motion}  &\multirow{2}{*}{Text}  &\multirow{2}{*}{Audio}  &\multirow{2}{*}{RGB}  &\multicolumn{2}{c}{Abstract}  &\multicolumn{3}{|c|}{Fine-grained}  &\multirow{2}{*}{Subj.}  &\multirow{2}{*}{Dur.}  &\multirow{2}{*}{FPS}  &\multirow{2}{*}{Lang.}  &\multirow{2}{*}{Repr.}  &\multirow{2}{*}{Env.}  &\multirow{2}{*}{Tech.} \\
\cline{6-10}
& & & & &\multicolumn{1}{c}{Act.}  &Emo.  &\multicolumn{1}{|c}{Exp.}   &\multicolumn{1}{c}{Pose}   &Scn.  & & & & & &
\\
\hline
BIWI \cite{fanelli20103}  &\checkmark  &-  &\checkmark  &-  &-  &L  &-  &-  &-  &14   &1.44h  &25   &EN   &Mesh   &\multirow{8}{*}{Lab}     &\multirow{8}{*}{3D}                  
\\
VOCASET \cite{cudeiro2019capture}  &\checkmark  &-  &\checkmark  &-  &-  &-  &-  &-  &-  &12  &0.5h &60   &EN   &Mesh       
\\
MeshTalk \cite{richard2021meshtalk}  &\checkmark  &-  &\checkmark  &-  &-  &-  &-  &-  &-  &250  &13h  &30   &EN   &Mesh                    
\\
Multiface \cite{wuu2022multiface}  &\checkmark  &-  &\checkmark  &\checkmark  &-  &-  &\checkmark  &-  &-  &13  &-  &30   &EN   &Mesh                    
\\
MMFace4D \cite{wu2023mmface4d}  &\checkmark  &-  &\checkmark  &-  &-  &L  &-  &-  &-  &431  &36h  &30   &CN   &Mesh            
\\
D3DFACS \cite{cosker2011facs}  &\checkmark  &-  &-  &\checkmark  &-  &-  &L  &-  &-  &10  &-  &60   &-   &Mesh            
\\
CoMA \cite{ranjan2018generating}  &\checkmark  &-  &-  &-  &-  &-  &L  &-  &-  &12  &-  &60   &-   &Mesh            
\\
4DFAB \cite{cheng20184dfab}  &\checkmark  &-  &-  &\checkmark  &-  &-  &L  &-  &-  &180  &-  &60   &-   &Mesh            
\\
\hline  
3D-ETF \cite{peng2023emotalk}   &\checkmark  &-  &\checkmark  &\checkmark  &-  &L  &-  &-  &-  &100+  &6.5h  &-   &EN   &Blendshape   &Mix   &Mono.
\\
MEAD-3D \cite{danvevcek2023emotional,he2023speech4mesh}   &\checkmark  &-  &\checkmark  &\checkmark  &-  &L  &-  &-  &-  &60  &38h  &30   &EN   &FLAME  &Lab   &Mono. 
\\
TEAD \cite{zhong2024expclip}   &-  &\checkmark  &-  &-  &-  &\checkmark  &-  &-  &\checkmark  &-  &-  &-  &-   &Blendshape  &-  &Gen.   
\\
TA-MEAD \cite{ma2023talkclip}   &-  &\checkmark  &\checkmark  &\checkmark  &-  &\checkmark  &\checkmark\kern-1.2ex\raisebox{0.7ex}{\rotatebox[origin=c]{125}{\textbf{--}}}   &-  &\checkmark  &60  &40h  &30   &EN   &-  &Lab  &-
\\
\rowcolor[gray]{.92}
\textbf{MMHead}   &\checkmark  &\checkmark  &\checkmark  &\checkmark  &\checkmark  &\checkmark  &\checkmark  &\checkmark  &\checkmark  &2$K$+  &49h  &25   &Mul.  &FLAME  &Mix  &Mono.           
\\
\bottomrule
\end{tabular}
}
\end{table*}

3D facial animation is becoming an increasingly popular topic in computer vision and multimedia due to its numerous applications in the multimedia field such as AR/VR content creation, games, and film production. The strong correlation between speech audio and facial movements makes it possible to automatically generate 3D facial motion (which can then be used to animate avatars) from audio, which will greatly simplify the animation production pipelines. 
Text, as another commonly used modality in human society, plays an important role in human-controlled AIGC. Recently, text guided image \cite{betker2023improving, rombach2022high}, video \cite{singer2022make, wang2023modelscope} and human motion \cite{tevet2022human, zhang2024motiondiffuse, guo2022generating} generation have achieved pleasing results with numerous multimedia applications. 
It would also be desirable to generate 3D facial motion from text descriptions. Further, it would be even more inspiring to achieve fine-grained 3D facial animation under multi-modal signal control. However, due to the lack of relevant datasets, fine-grained multi-modal 3D facial animation has been rarely explored.

In the field of 3D facial animation, impressive progress has been made in audio-driven 3D facial animation \cite{cudeiro2019capture,xing2023codetalker,peng2023selftalk,wu2023speech,thambiraja2023imitator,stan2023facediffuser,aneja2024facetalk}, making it possible to generate 3D facial motion sequences with synchronized lip movements based on the input talking audio. Moreover, some works \cite{peng2023emotalk,danvevcek2023emotional,wang20213d} have incorporated emotion labels to achieve explicit control over talking emotions. Recently, \cite{ma2023talkclip,zhong2024expclip,zhao2024media2face} tried to use text descriptions in natural language rather than emotion labels to guide the facial animation. 
However, they only consider textual descriptions of emotions during talking, while overlooking the actions other than talking, as well as text descriptions of fine-grained facial and head movements.
Such limitations seriously affect the convenience and flexibility of 3D face animation in multimedia applications.
A major reason for these limitations is the lack of open-source multi-modal 3D facial animation datasets with fine-grained text annotations, as shown in Tab. \ref{tab:tab1}.

To alleviate the scarcity of multi-modal 3D facial animation datasets, we present MMHead, a large and diverse multi-modal 3D facial animation dataset with rich hierarchical text annotations. To the best of our knowledge, MMHead is the first 3D facial animation dataset with both abstract and fine-grained text annotations, including abstract action and emotion descriptions, fine-grained facial and head movement (\textit{i.e.}, expression and head pose) descriptions, as well as possible scenarios that may cause the person's emotion (see Fig. \ref{fig:teaser} and Tab. \ref{tab:tab1}).
Concretely, we build our 3D facial animation dataset from 2D portrait video datasets \cite{zhu2022celebv,yu2023celebv,wuu2022multiface,wang2020mead,livingstone2018ryerson}, considering the huge amount of 2D videos and satisfactory monocular 3D face reconstruction results \cite{danvevcek2022emoca,DECA:Siggraph2021,filntisis2022visual}. An automatic pipeline is also proposed to reconstruct 3D facial motion sequences from monocular videos, and then obtain abstract and fine-grained text annotations for these facial motions. The data construction pipeline can be easily scaled up to obtain larger datasets.
To be specific, we first integrate and filter five public 2D portrait video datasets, \textit{i.e.}, CelebV-HQ \cite{zhu2022celebv}, CelebV-Text \cite{yu2023celebv}, MEAD \cite{wang2020mead}, RAVDESS \cite{livingstone2018ryerson}, Multiface \cite{wuu2022multiface}, to achieve richer facial actions and emotions. Then, we reconstruct 3D facial motion represented by FLAME parameters \cite{li2017learning} via a state-of-the-art emotion-preserving monocular 3D reconstruction method named EMOCA \cite{danvevcek2022emoca,DECA:Siggraph2021,filntisis2022visual}. Optimization and manual screening are then performed to obtain final 3D facial motion sequences that can achieve comparable precision compared to the lab-collected 3D facial animation datasets. 
As for text annotation, we explore the annotation capabilities of the large language model through well-designed prompts. Specifically, the action and emotion labels of the portrait video datasets, per frame activated facial Action Units (AU) \cite{luo2022learning,wang2023agentavatar} and head poses are conditionally combined with five different prompts to feed into ChatGPT \cite{chatgpt} to obtain natural text descriptions of abstract action, abstract emotion, detailed expression, detailed head pose, and emotion scenarios, respectively. 

Along with the proposed multi-modal 3D facial animation dataset, we benchmark two novel tasks: 1) text-induced 3D talking head animation (Benchmark I), which aims at generating 3D facial animations according to both speech audio and text instructions; 2) text-to-3D facial motion generation (Benchmark II), \textit{i.e.}, generating 3D facial motion sequences based on the given text descriptions. We define these tasks as non-deterministic generation tasks \cite{yang2023probabilistic,zhang2023generating} rather than regression tasks \cite{cudeiro2019capture,fan2022faceformer,xing2023codetalker} to achieve robust and diverse 3D facial motion generation. 
Moreover, we propose a simple but efficient method named MM2Face to unify the multi-modal information and generate diverse and plausible 3D facial motions, which first utilizes a VQ-VAE \cite{van2017neural,esser2021taming} to compress the 3D facial motions into a discrete codebook, and then generate plausible results in a constrained space by sampling the codebook in an autoregressive fashion through a transformer \cite{vaswani2017attention} equipped with specially designed attention maps.
Based on MM2Face, we further explore the fusion and injection strategies of text and audio modalities to provide insight for future research.

In summary, our main contributions are:
\begin{itemize}
\setlength\itemsep{.8em}
    \item We propose the first multi-modal 3D facial animation dataset, MMHead, which contains rich hierarchical text annotations including abstract action and emotion descriptions, fine-grained facial and head movements descriptions, and possible emotion scenarios.
    \item We benchmark two new tasks: text-induced 3D talking head animation and text-to-3D facial motion generation to promote future research on multi-modal 3D facial animation.
    \item We propose a VQ-VAE-based method to unify the multi-modal information and generate diverse and plausible 3D facial motions. The proposed framework achieves competitive results on both benchmarks.
\end{itemize}

\section{Related Work}
\noindent\textbf{Audio-driven 3D Facial Animation.} 
Audio-driven 3D facial animation aims at generating 3D facial motion sequences according to the input speech audio. In recent years, it has attracted much attention due to its potential use in virtual avatar animation \cite{zheng2022avatar,wu2023ganhead,wang2023versatile}, film, and game production.
Audio-driven 3D facial animation methods can be roughly divided into two categories: rule-based methods \cite{cohen2001animated,taylor2012dynamic,xu2013practical,edwards2016jali} and learning-based methods \cite{cudeiro2019capture,richard2021meshtalk,fan2022faceformer,xing2023codetalker,peng2023emotalk,peng2023selftalk,wu2023speech,danvevcek2023emotional,stan2023facediffuser,aneja2024facetalk,sun2023diffposetalk,thambiraja20233diface}.  
Rule-based methods typically establish complex mapping rules between pronunciations and lip motions. This kind of method makes it easy to ensure the accuracy of the lip motion, however, requires a lot of manual effort and cannot generate the motion of the entire head.
Learning-based methods solve the 3D facial animation problem in a data-driven way, and have made impressive progress in recent years \cite{cudeiro2019capture,richard2021meshtalk,fan2022faceformer,xing2023codetalker,peng2023selftalk,wu2023speech,stan2023facediffuser,aneja2024facetalk,sun2023diffposetalk,thambiraja20233diface}. Furthermore, EmoTalk \cite{peng2023emotalk} and EMOTE \cite{danvevcek2023emotional} introduce emotion labels and achieve explicit control over talking emotions. ExpCLIP \cite{zhong2024expclip} and Media2Face \cite{zhao2024media2face} go one step further and use emotion text descriptions rather than labels to control the emotion of audio-driven 3D facial animation. However, they only consider the textual control of emotions during talking, while ignoring the fine-grained control over facial and head movements.
In addition, text-to-3D facial motion generation without input speech audio has never been explored except for expression generation \cite{otberdout2022sparse,zou20234d}, which affects the convenience and flexibility of 3D face animation in multimedia applications.
To fill this gap and promote the research on fine-grained multi-modal 3D facial animation, we present the first large-scale multi-modal 3D facial animation dataset with rich hierarchical text descriptions.

\noindent\textbf{Text-guided Generation.} Incorporating text descriptions into generative models has been a popular topic in AIGC areas such as text-to-image/video/motion generation. In the common setting of the text-to-image task, generative models \cite{goodfellow2020generative, ho2020denoising, van2017neural, kingma2013auto} are widely used to capture the multi-modal distributions of images, and can generate diverse and plausible images compared to the regression models. Early works \cite{xia2021tedigan, kang2023scaling, tevet2022motionclip, guo2020action2motion, zhai2023language} often use GANs or VAEs with well-designed networks for text-guided generation. Recently, denoising diffusion models \cite{singer2022make, zhang2023show, tevet2022human, rombach2022high, chen2023executing, zhang2024motiondiffuse, yang2023synthesizing} and VQ-VAE have been explored with superior performance. VQ-VAE is first explored in the text-to-image task with numerous works \cite{esser2021taming, yu2022scaling, chang2023muse}. In recent years, lots of works \cite{zhang2023t2m, guo2022tm2t, guo2023momask, jiang2024motiongpt, pinyoanuntapong2023mmm, xu2023inter, liang2024intergen, lin2024motion, petrovich2023tmr} also focus on text-to-3D human motion generation and achieve promising results. Despite the impressive progress of text-guided generation in numerous areas, text-guided 3D facial animation has been rarely explored before. 
To this end, we explore the text-to-3D facial motion generation task based on the proposed MMHead dataset in this paper.

\noindent\textbf{3D Facial Animation Datasets.} 
3D facial animation datasets play an important role in both research and multimedia applications, which can be roughly divided into audio-driven 3D facial animation datasets and 3D facial expression datasets. 
Audio-driven 3D facial animation datasets \cite{richard2021meshtalk, cudeiro2019capture, fanelli20103} collect dynamic 3D facial motions with synchronized speech audios and gradually seek to construct larger, higher quality datasets. However, due to the high cost of collecting dynamic 3D data, such datasets are hard to scale up, which has prompted numerous works \cite{peng2023emotalk,sun2023diffposetalk,danvevcek2023emotional,he2023speech4mesh,wu2023singinghead} to reconstruct 3D facial motion from portrait video datasets.
Recently, other control signals have begun to emerge in audio-driven 3D facial animation datasets. EmoTalk \cite{peng2023emotalk} and EMOTE \cite{danvevcek2023emotional} present datasets with emotion labels. ExpCLIP \cite{zhong2024expclip} proposes a text-expression alignment dataset, however, it only contains static facial expressions and ignores facial actions other than talking as well as the corresponding text descriptions.
As for 3D facial expression datasets, some works \cite{cosker2011facs,zhang2014bp4d,zhang2016multimodal,ranjan2018generating,cheng20184dfab,wuu2022multiface,pan2024renderme} capture someone's 3D face sequences when asking him/her to make a given expression, which lack of fine-grained text annotations and other facial actions other than the specified expressions. Different from existing 3D facial animation datasets, our MMHead dataset contains both speech audio and rich hierarchical text annotations of diverse facial actions.

\section{MMHead Dataset}
\label{sec:dataset}

\begin{figure*}[t]
\centering
\includegraphics[width=\linewidth]{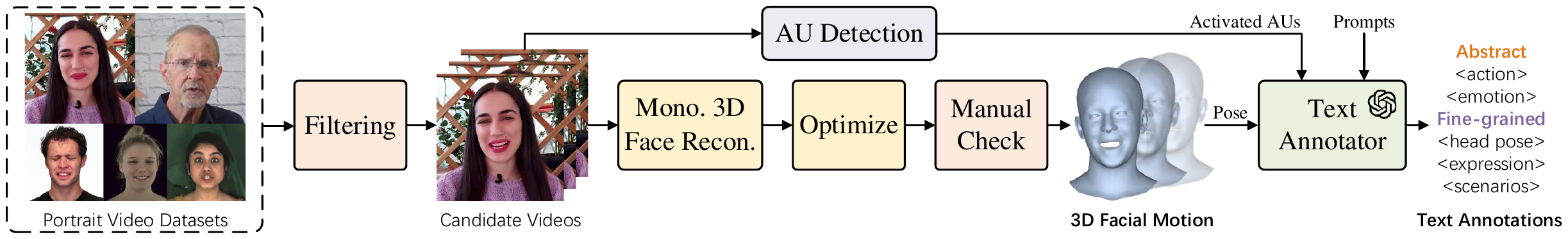}
\caption{\textbf{Dataset construction pipeline.} We first integrate five portrait video datasets and filter the data to obtain the candidate videos for constructing our 3D facial animation dataset. Then, high-precision 3D facial motion sequences are obtained from candidate videos through monocular 3D face reconstruction, FLAME parameter optimization, and manual screening in turn. Finally, we utilize ChatGPT with well-designed prompts to obtain abstract and fine-grained text annotations.
}
\label{fig:pipeline}
\end{figure*}

Considering the absence of a fine-grained multi-modal 3D facial animation dataset, we present MMHead, a large-scale dataset with both speech audio and fine-grained text annotations. 
As shown in Fig. \ref{fig:pipeline}, we first collect a portrait video candidate set from five publicly available portrait video datasets (Sec. \ref{sec:3.1}), and then reconstruct 3D facial motion from this candidate set through monocular 3D face reconstruction technologies \cite{danvevcek2022emoca,DECA:Siggraph2021,filntisis2022visual, kingma2014adam} followed by optimization and manual check (Sec. \ref{sec:3.2} part 1). 
In addition, we propose an automatic text annotation pipeline to obtain rich text annotations including abstract action and emotion descriptions, fine-grained facial and head movements descriptions, and three possible scenarios that may cause the person's emotion (Sec. \ref{sec:3.2} part 2).
The statistics analysis of the MMHead dataset, and dataset partitioning for the two benchmarks are detailed in Sec. \ref{sec:3.3}.

\subsection{Data Collection}
\label{sec:3.1}
\noindent\textbf{Dataset Integration.} 
We integrate five commonly used portrait video datasets with rich facial action and emotion to obtain the candidate video set for our 3D facial motion. 
(1) \textbf{CelebV-HQ} \cite{zhu2022celebv}, a large-scale and high-quality in-the-wild face video dataset with manually labeled facial action and emotion labels. We remove videos with actions that are hard to see on 3D head, including "kiss", "listen to music", "play instrument", "smoke", and "whisper", and reuse the remaining action and emotion labels in our automatic text annotation pipeline for abstract text descriptions. 
(2) \textbf{CelebV-Text} \cite{yu2023celebv}, a large-scale and high-quality in-the-wild dataset with facial text-video pairs. Although it has text descriptions for dynamic actions and emotions, it often contains useless descriptions and emotion annotations that change too frequently. To this end, we remove those videos with emotion changes more than four times, and only use the action and emotion labels of the CelebV-Text dataset. Similar to CelebV-HQ, we remove videos with inconspicuous action labels, here "squint" action is also removed.
(3) \textbf{MEAD} \cite{wang2020mead}, a lab-collected high-quality multi-view emotional talking head video dataset, in which each speaker speaks with 8 emotions in three intensity levels. Here we only use the frontal videos. To balance talking with other actions as well as achieve more obvious emotion labels, we only use the level-3 data for each emotion in the dataset.
(4) \textbf{RAVDESS} \cite{livingstone2018ryerson}, a lab-collected high-quality emotional talking and singing video dataset with 8 emotions. Here we use all of its data.
(5) \textbf{Multiface} \cite{wuu2022multiface}, a lab-collected multi-view facial animation dataset with various facial expressions and several talking sequences. Here we only use the frontal videos of recognizable expressions.

\noindent\textbf{Video Filtering.} 
We first unify all the portrait videos we got above into 25 FPS, and then remove videos with more than 200 frames.
Moreover, we filter out videos with poor audio quality, since the synchronization between audio and facial motion, especially mouth motion, is of great importance, however, out-of-sync sound and picture, excessive background noise or background music, and multiple people speaking together in one video are common in in-the-wild videos.
We follow VoxCeleb \cite{chung2018voxceleb2} and utilize SyncNet \cite{chung2017out}, a pre-trained model for determining the audio-video synchronization, to get the confidence score of each in-the-wild portrait video, and then filter out videos with confidence score lower than a specified threshold.

\begin{figure}[h]
\centering
\includegraphics[width=\linewidth]{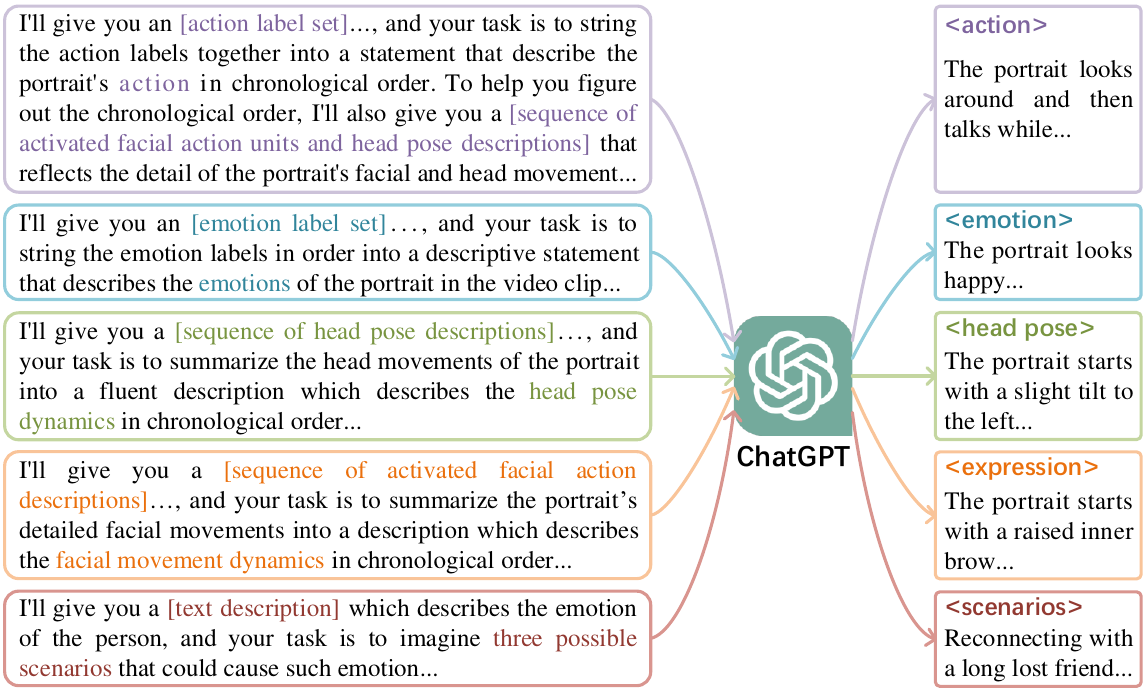}
\caption{\textbf{Text annotation pipeline with well-designed prompts.} We use different prompts and input information to annotate each data with five types of text descriptions separately by ChatGPT.}
\label{fig:annotator}
\end{figure}

\subsection{Automatic Annotation Pipeline}
\label{sec:3.2}

\noindent\textbf{3D Facial Motion Reconstruction.}
We utilize sequences of expression (50 dimensions) and pose (6 dimensions) parameters of a commonly used parametric head model, FLAME \cite{li2017learning}, to represent 3D facial motion. To reconstruct FLAME parameters from portrait videos, we select a state-of-the-art emotion-preserving monocular 3D face reconstruction method, EMOCA \cite{danvevcek2022emoca,DECA:Siggraph2021,filntisis2022visual}, to estimate FLAME parameters for each frame. 
However, due to the shaking of the people and complex background in the videos, frame-by-frame reconstruction results may exhibit instability, such as incorrect shapes and jittery motions. To solve this problem, we first sieve out the outliers of the estimated FLAME parameters through Box Plot and replace them with the values of neighboring points, and then smooth the expression and pose parameters via per-video optimization. All 3D facial motion sequences are manually checked to guarantee quality. 
For more details, please refer to the Sup. Mat.

\noindent\textbf{Text Annotation.}
We annotate each 3D facial motion with rich text descriptions including abstract action and emotion descriptions, fine-grained facial and head movements descriptions, and three possible scenarios that may cause such emotion. These five types of text descriptions are annotated separately as illustrated in Fig. \ref{fig:annotator}. For full content and format of the well-designed prompts, please refer to the Sup. Mat.
(1) \textbf{Abstract action}. 
Since the lab-collected part of our candidate video set only contains one action in "talking", "singing" and "making a facial expression", we predefine seven text descriptions for each of these three actions, and then randomly select one of the seven predefined texts as the abstract action text description for each 3D facial motion from lab-collected videos.
As for 3D facial motions from in-the-wild videos, multiple actions may occur in one motion sequence and we only have the action labels that are not strictly in chronological order, so we feed ChatGPT \cite{chatgpt} additional information including per frame activated AU \cite{luo2022learning,wang2023agentavatar} and head pose labels with their intensity values, and let ChatGPT deduce the chronological order of the action labels, and then give an abstract action text description according to the chronological order. 
(2) \textbf{Abstract emotion}.
We let ChatGPT string the given emotion labels (borrow from original video datasets) into a natural text description which will serve as our abstract emotion annotation.
(3) \textbf{Fine-grained facial movements (\textit{i.e.}, expression)}.
We find that letting ChatGPT annotate detailed facial and head movements separately can achieve better performance than annotating them together. Here we feed ChatGPT per frame-activated AU labels together with their intensity values, and ask ChatGPT to summarize them in chronological order to obtain the fine-grained facial movements text descriptions.
(4) \textbf{Fine-grained head movements (\textit{i.e.}, head pose)}.
We define seven types of head pose state labels according to the neck rotation vector of the FLAME \cite{li2017learning} model, and feed ChatGPT per frame head pose labels with corresponding rotation angles to obtain the fine-grained head movements text descriptions. 
(5) \textbf{Scenarios}.
We finally ask ChatGPT to imagine three possible scenarios that may cause the person to experience the emotions annotated in (2).
Note that these hierarchical text descriptions can be automatically annotated in batches, which can be easily extended to larger datasets.

\subsection{Dataset Analysis} 
\label{sec:3.3}

\noindent\textbf{Data Statistics.} 
As described above, we construct MMHead from five online portrait video datasets and end up with 35903 3D facial motion sequences up to 49 hours in total. Specifically, there are 10505, 11128, 11313, 2452, 505 motion sequences from CelebV-HQ \cite{zhu2022celebv}, CelebV-Text \cite{yu2023celebv}, MEAD \cite{wang2020mead}, RAVDESS \cite{livingstone2018ryerson}, Multiface \cite{wuu2022multiface}, respectively.
Each 3D facial motion is annotated with five types of text annotations including abstract action and emotion, fine-grained facial and head movements, and three possible scenarios.
For more details about the motion duration, action and emotion categories statistics, please refer to the Sup. Mat.

\noindent\textbf{Dataset partitioning for Two Benchmarks.} 
We benchmark two tasks based on the MMHead dataset. For text-induced talking head animation (Benchmark I), we use a total of 28466 3D facial motion sequences related to "talk", "sing", and "read". For text-to-3D facial motion generation (Benchmark II), we use 7937 sequences in total, including 505 sequences with various facial expressions from the Multiface dataset, 250 talking, and 250 singing sequences randomly sampled from the data used in benchmark I, and a total of 6932 sequences for all actions other than "talk", "sing", and "read".

\section{MM2Face Method}

\begin{figure*}[t]
\centering
\includegraphics[width=\linewidth]{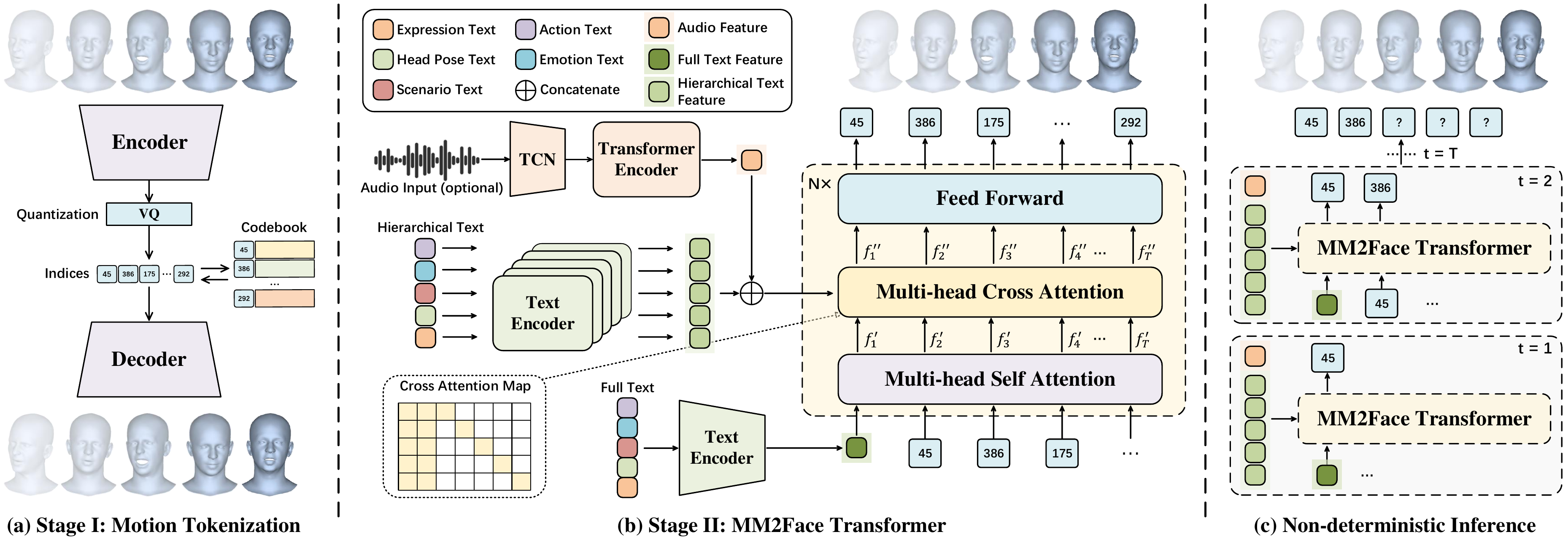}
\caption{\textbf{Overview of our MM2Face framework.} In stage I, we utilize a VQ-VAE $\mathcal{V}$ to tokenize the FLAME facial motions $\boldsymbol{F}_{1:T}$ to a sequence of motion tokens. Then in stage II, we utilize a causal auto-regressive transformer MM2Face $\mathcal{G}$ to generate discrete motion tokens $\boldsymbol{\Tilde{F}}$ sequentially from audio and text inputs.}
\vspace{-1mm}
\label{fig:MM2Face}
\end{figure*}

We explore a simple but efficient two-stage method MM2Face that can handle both newly proposed text-induced 3D talking head animation and text-to-3D facial motion generation tasks. MM2Face can synthesize diverse and plausible 3D facial motions given a speech audio sequence (optional) and a fine-grained text description.

Our MM2Face framework consisting of two stages is illustrated in Fig. \ref{fig:MM2Face}. In the first stage (Sec. \ref{face_tokenization}), given a 3D facial motion sequence, we utilize a motion tokenizer to extract discrete and compact 3D facial motion tokens. In the second stage (Sec. \ref{stage2_method}), we conduct an auto-regressive token modeling task in discrete token space using our MM2Face transformer \cite{vaswani2017attention}. During inference, MM2Face iteratively predicts the likelihood of motion tokens in each step, enabling non-deterministic motion generation.

\subsection{Facial Motion Tokenization}
\label{face_tokenization}

In the first stage, we utilize VQ-VAE to extract meaningful discrete facial motion representations. We utilize a 1D CNN-based network \cite{he2016deep} to train a facial motion VQ-VAE $\mathcal{V}$ on diverse 3D facial motion data, which consists of a motion encoder $\mathcal{E}$, a vector quantization network $\mathcal{Q}$, a codebook $\mathcal{B}$, a motion decoder $\mathcal{D}$. Given a motion sequence $\boldsymbol{F}_{1:T} = (\boldsymbol{F}_1, \boldsymbol{F}_2, ..., \boldsymbol{F}_T) \in \mathbb{R}^{T \times d}$, where $T$ denotes the motion frame number and $d$ denotes the dimension of FLAME parameters. The encoder $\mathcal{E}$ extract meaningful feature vectors $\boldsymbol{z} = (\boldsymbol{z}_1, \boldsymbol{z}_2, ..., \boldsymbol{z}_{T^{'}})$ from $F$, where $T^{'} = T/r$, $r$ is the temporal downsampling rate ($r = 4$ in our paper). Then the feature vector $\boldsymbol{z}$ is processed by the vector quantization network to get quantized features $\boldsymbol{\hat{z}} = (\boldsymbol{\hat{z}}_1, ..., \boldsymbol{\hat{z}}_{T^{'}})$ via searching nearest neighbors in the codebook $\mathcal{B}$, which can be written as:
\begin{equation}
\boldsymbol{\hat{z}}_j = argmin_{b_k \in \mathcal{B}} ||\boldsymbol{z}_{j} - \boldsymbol{b}_{k}||.
\end{equation}
Finally, the quantized feature $\boldsymbol{\hat{z}}$ is fed into the motion decoder $\mathcal{D}$ to reconstruct human motions $\boldsymbol{F}$. Through the process, we can convert facial motion data to discrete features and their corresponding codebook indices $I$, \textit{i.e.}, motion tokens.

\noindent \textbf{Training Loss.} We adopt the motion reconstruction loss, embedding commitment loss for training. The reconstruction loss is:
\begin{equation}
\mathcal{L}_{rec} = \mathcal{L}_{mse}(\boldsymbol{F}_{1:T}, \boldsymbol{\hat{F}}_{1:T}) + \alpha \mathcal{L}_{mse}(V(\boldsymbol{F}_{1:T}), V(\boldsymbol{\hat{F}}_{1:T})),
\end{equation}
where $\mathcal{L}_{mse}$ denotes the MSE loss function, $V(\boldsymbol{F}_{1:T}) = \boldsymbol{F}_{2:T} - \boldsymbol{F}_{1:T-1}$ denotes the velocity of $\boldsymbol{F}$. $\alpha$ is a hyper-parameter.

The commitment loss is:
\begin{equation}
\mathcal{L}_{commit} = ||sg(\boldsymbol{\hat{z}} ) - \boldsymbol{z}|| + \beta ||sg(\boldsymbol{z} ) - \boldsymbol{\hat{z}}||,
\end{equation}
where $\beta$ is a hyper-parameter, $sg$ denotes the stop gradient operation.
Finally, we combine these losses $\mathcal{L}_{vq} = \mathcal{L}_{rec} + \mathcal{L}_{commit}$ to optimize the VQ-VAE network.

\subsection{Auto-regressive Token Modeling}
\label{stage2_method}

After converting all the facial motions to discrete tokens, auto-regressive token modeling is conducted to train our MM2Face transformer model $\mathcal{G}$. The MM2Face transformer, as shown in Fig. \ref{fig:MM2Face}, consists of an audio encoder $\mathcal{A}$, a full text encoder $\mathcal{T}_{full}$, hierarchical text encoders $\mathcal{T}_{h1:h5}$ and a MM2Face transformer $\mathcal{C}$. 

Specifically, the audio $\boldsymbol{A}_{1:T}$ is fed into $\mathcal{A}$ to extract dense audio features, we concatenate all fine-grained text descriptions into a full text and feed it into $\mathcal{T}_{full}$ to obtain full text feature, fine-grained texts are separately fed into hierarchical text encoders $\mathcal{T}_{h1:h5}$ to obtain hierarchical text features, then audio features and hierarchical text features are concatenated together as cross attention inputs. The MM2Face transformer $\mathcal{C}$ takes the ground truth motion tokens, audios, full and hierarchical text features as inputs and sequentially predicts the generated motion tokens $\Tilde{I}$. Then $\Tilde{I}$ can be decoded to final generated facial motions $\boldsymbol{\Tilde{F}}$.

\noindent \textbf{Architecture.} Our audio encoder $\mathcal{A}$ is a pre-trained state-of-the-art speech model wav2vec 2.0 \cite{baevski2020wav2vec}, which is composed of a temporal convolution-based network and a multi-layer transformer encoder. $\mathcal{A}$ takes the audio inputs $\boldsymbol{A}_{1:T}$ and output the audio feature. The text encoder $\mathcal{T}$ is a pre-trained distilbert \cite{sanh2019distilbert} model. The MM2Face transformer $\mathcal{C}$ is a multi-layer transformer decoder. Each layer of $\mathcal{C}$ is composed of normalization layers, a self-attention layer, a feed-forward layer, and a cross-attention layer.

\noindent \textbf{Biased Cross-attention.} Giving the concatenated hierarchical text features and audio features as inputs for cross-attention calculation, we specifically design a biased cross-attention mask $mask$ in Fig. \ref{fig:MM2Face} to align the motion tokens with inputs:
\begin{equation}
Att = Softmax\left ( \frac{Q^F(K)^T\times mask}{\sqrt{d_k} }  \right ) ,
\end{equation}
where $K \in R^{T \times d_k}$ is the key of the concatenated hierarchical text features and audio features.

\noindent \textbf{Training Loss.} During training, we adopt the teacher-forcing strategy and directly maximize the log-likelihood of the facial motion token representations:
\begin{equation}
L_{MM2Face} = E_{x\sim p(x)}[-log p(x|c)].
\end{equation}


\noindent \textbf{Diverse Probabilistic Facial Motion Inference.} Different from previous deterministic methods \cite{fan2022faceformer, xing2023codetalker}, our method can generate diverse facial motions with various head poses and expressions given input audio and text descriptions. Specifically, MM2Face iteratively predicts the likelihood of motion tokens in each step, then can stochastically generate the motion token in each step through predicted likelihood. 




\section{Experiments}

In this section, we sequentially introduce our proposed two benchmarks, \textit{i.e.}, text-induced 3D talking head animation and text-to-3D facial motion generation. Concretely, we extensively evaluate the previous methods and our MM2Face method on these benchmarks.



\begin{table*}[t]
    \centering
    \caption{\textbf{Experimental results for benchmark I: text-induced 3D talking head animation.} The best and runner-up performances are bold and underlined, respectively.}
    \resizebox{\linewidth}{!}{

    \begin{tabular}{l c c c c c c c c}
    \toprule
    \multirow{2}{*}{Methods}  & \multicolumn{3}{c}{Text R-Precision $\uparrow$} & \multirow{2}{*}{FID $\downarrow$} & \multirow{2}{*}{Audio-Match $\downarrow$} & \multirow{2}{*}{Diversity $\rightarrow$} & \multirow{2}{*}{LVE (mm) $\downarrow$} & \multirow{2}{*}{FVE (mm) $\downarrow$} \\

    \cline{2-4}
    ~ & Top-1 & Top-2 & Top-3 \\
    
    \midrule

        Real facial motion & 0.843 & 0.941 & 0.970 & 0 & 31.70 & 50.84 & 0 & 0 \\
        Our VQ-VAE \small{(Recons.)} & 0.777 & 0.892 & 0.934 & 14.62 & 35.03 & 50.37 & 2.855 & 0.807  \\ 
    \midrule

        FaceFormer \cite{fan2022faceformer} & 0.353 & 0.477 & 0.548  & 667.7 & 34.53 & 44.17 & 6.790 & \underline{1.692} \\

        CodeTalker \cite{xing2023codetalker} & 0.303 & 0.409  & 0.471  & 579.8  & 43.32  & 47.14  & 9.057  & 2.097  \\

        SelfTalk \cite{peng2023selftalk} & 0.286  & 0.399 & 0.478 & 708.4   & \underline{33.19}  &  43.23  & \underline{6.766}  & \textbf{1.680} \\
        
        Imitator \cite{thambiraja2023imitator}  & 0.328 & 0.471 & 0.555  & 763.3 & 37.11 & 45.31 & 6.792 & 1.693 \\ 

        FaceDiffuser \cite{stan2023facediffuser}  & \underline{0.703}  & \underline{0.853}  & \underline{0.905} & 72.03 &  \textbf{32.78}   & \textbf{50.53} &  7.905  & 1.903 \\

    \midrule
        Our MM2Face (w/o Text) & 0.404  & 0.520  & 0.584  & \underline{50.51}  & 34.66  & \underline{50.37} & 7.053  & 1.804 \\

        \rowcolor[gray]{.92}
        \textbf{Our MM2Face} & \textbf{0.718} & \textbf{0.854} & \textbf{0.909} & \textbf{41.19} &  34.57 & 50.09 & \textbf{6.736} & \underline{1.692} \\

    \bottomrule
    \end{tabular} 
    }
    \label{fid_table}

\end{table*}

\subsection{Text-induced 3D Talking Head Animation}

The provided fine-grained text annotations with corresponding audio and 3D facial motions lead to a new task, \textit{i.e.}, text-induced 3D talking head animation. Since there are no publicly available methods focusing on text-induced 3D talking head animation, we simply add full text feature to existing audio-driven methods as baseline methods. We select five state-of-the-art audio-driven 3D facial animation methods, \textit{i.e.}, FaceFormer \cite{fan2022faceformer}, CodeTalker \cite{xing2023codetalker}, SelfTalk \cite{peng2023selftalk}, Imitator \cite{thambiraja2023imitator}, and FaceDiffuser \cite{stan2023facediffuser}. Concretely, the full text feature is extracted through a text encoder same as the proposed MM2Face method, and injected through self-attention for transformer-based methods \cite{fan2022faceformer,xing2023codetalker,peng2023selftalk,thambiraja2023imitator} and directly concatenated with the audio feature for diffusion-based method \cite{stan2023facediffuser}.

\noindent\textbf{Experiment setup.} We adopt the common protocol to split our dataset into training, validation, and test sets
with a ratio of 0.8, 0.05, and 0.15. We select the 56 dimensions FLAME parameters as the input of our method and evaluate the generation performance on generated mesh sequences for fair comparison with other mesh-based methods.



\noindent\textbf{Metrics.} Previous approaches \cite{fan2022faceformer, xing2023codetalker} treat this task as a deterministic prediction task instead of a generative task, hence only the accuracy-related metrics are calculated. To evaluate the diversity and plausibility of generated motions, we follow the previous text to human motion task \cite{guo2022generating, guo2022tm2t} and select Frechet Inception Distance (FID), text R-Precision, audio matching score, diversity and lip vertices error \cite{fan2022faceformer} as our evaluation metrics: (1) \textbf{Frechet Inception Distance (FID)}, which is the distribution distance between the extracted features of generated motion and real motion by our pre-trained text-motion retrieval network, the network is trained following TMR \cite{petrovich2023tmr} which is a state-of-the-art retrieval method in 3D human motion area. (2) \textbf{Text R-Precision}: we calculate the top-1 and top-3 text to motion retrieval accuracy as reported metrics. We specifically compute them by ranking the Euclidean distance between the facial motion and text embeddings in a batch of 32 motion-text pairs. (3) \textbf{Audio Matching Score} is the average Euclidean distances between each gt audio feature and the generated motion feature extracted by our audio-motion retrieval network trained by the method mentioned in the supplementary materials. (4) \textbf{Diversity}. We randomly select 300 motion pairs from generated motions. Then We extract motion features and compute the average Euclidean distances of each motion pair to compute motion diversity in the test set. (5) \textbf{Lip Vertex Error (LVE)} \cite{richard2021meshtalk} calculates the maximum L2 error across all lip vertices for each frame. (6) \textbf{Face Vertex Error (FVE)} calculates the average L2 error across all face region vertices for each frame.



\noindent\textbf{Quantitative Evaluation.} The quantitative results are summarized in Tab. \ref{fid_table} on MMHead dataset. For motion reconstruction, our VQ-VAE achieves considerable reconstruction performance against real facial motion, demonstrating excellent discrete representations. For motion generation, we can observe that our MM2Face achieves the best performance against other state-of-the-art methods in almost all metrics. Concretely, the excellent Text R-Precision accuracy demonstrates that MM2Face is able to synthesize plausible 3D facial motions while correctly following the fine-grained text descriptions. The excellent audio matching score, LVE, and FVE demonstrate MM2Face's ability for precise motion generation. The FID score also indicates the generative distribution modeling performance of our MM2Face method. 



\begin{figure}
\centering
\includegraphics[width=\linewidth]{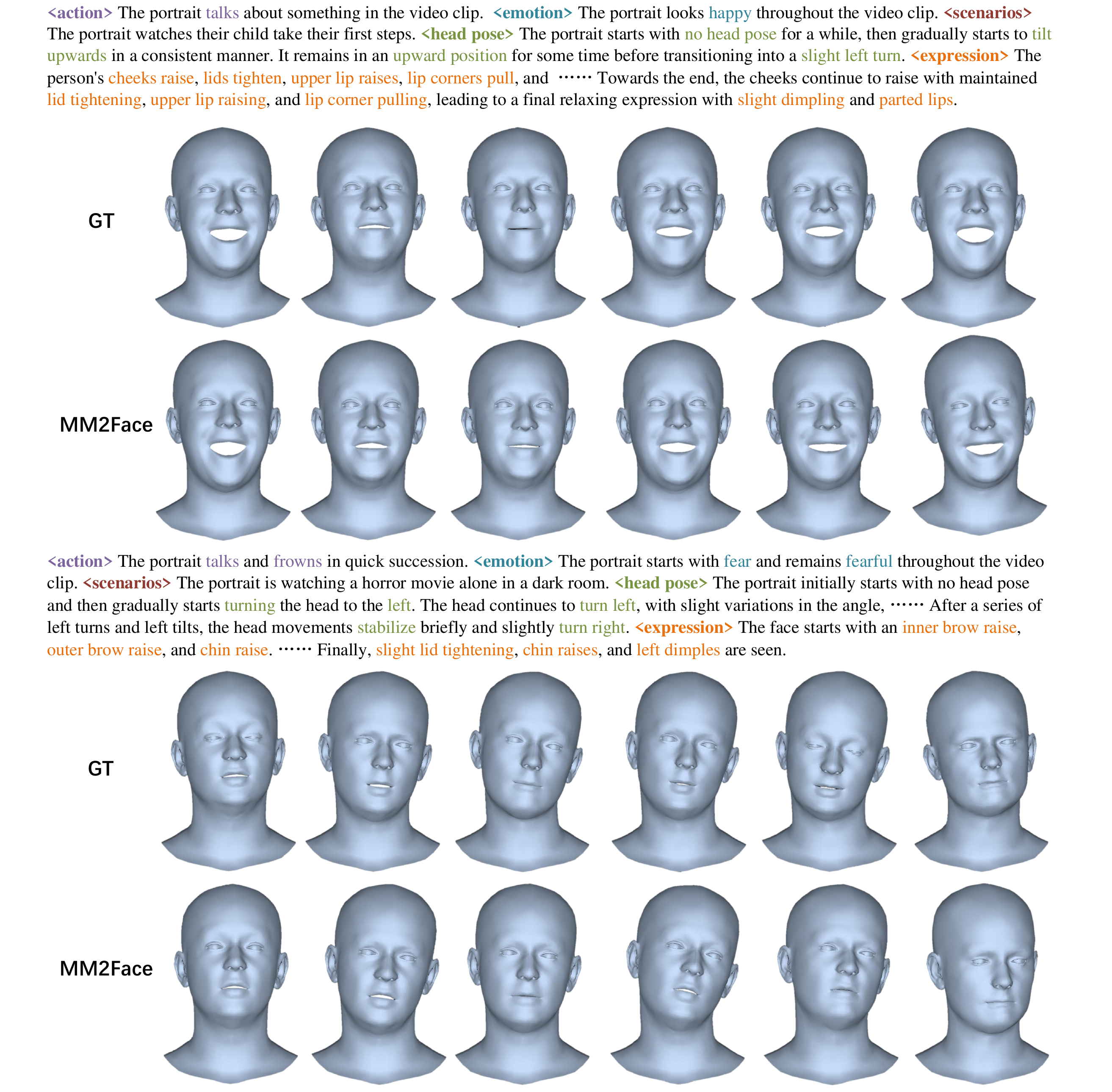}
\caption{Qualitative results of MM2Face on benchmark I: text-induced 3D talking head animation.}
\label{fig:talk_results}
\end{figure}

\noindent\textbf{Qualitative Evaluation.} We also provide the qualitative results in Fig. \ref{fig:talk_results}. Our approach exhibits superior text and face alignment with accurate audio-visual correspondence. More qualitative results can be seen in supplementary material.

\begin{table}
\centering
\caption{\textbf{Ablation study} of different multi-modal information fusion strategies in MM2Face. ‘FT', ‘HT', and ‘A' denote full text, hierarchical text, and audio, respectively.}
\resizebox{\linewidth}{!}{ 
\begin{tabular}{cc|ccc}
\toprule
Self-atten.  & Cross-atten.  & Top1-3 R-pre\,$\uparrow$ & FID\,$\downarrow$ & LVE\,$\downarrow$ \\ \hline
FT &    A        & [0.610, 0.776, 0.840]   &  45.68    &    6.985            \\
FT, HT &    A         & [0.714, 0.851, 0.906]   &  40.65     &     6.742            \\

FT,  HT,  A &   -      &  [0.656, 0.798, 0.864]  &   72.65   &   8.073         \\
  - &   FT,  HT,  A     &  [0.699, 0.846, 0.908]   &  35.57     &    6.824 \\
\rowcolor[gray]{.92}
FT &  HT,  A         &  [0.718, 0.854, 0.909] &  41.19   &  6.736              \\
\bottomrule
\end{tabular}
}
\label{table:ablation}
\end{table}

\noindent\textbf{Ablation Study.} Since our MM2Face method focuses on the multi-modal generation of facial motions, we perform an ablation study of various modality fusion strategies. Considering self-attention and cross-attention are the two most popular fusion mechanisms, we conducted a detailed study. The experiment results are summarized in Table. \ref{table:ablation}. First, we can observe that merely utilizing full text results in poor generation performance, demonstrating the necessity of separating fine-grained texts. Second, fusing all text features and audio features through the self-attention mechanism also produces bad results, it is because self-attention is difficult to model the dense alignment between generated motion tokens and corresponding audio. Third, we can see that the remaining ablations produce slightly similar results but full text self-attention, hierarchical text, and audio cross-attention yield the best R-precision, hence our MM2Face method adopts this architecture. Besides, we also explore the role of text inputs in the last two lines of Table. \ref{fid_table}, which demonstrates the effectiveness of text descriptions in controlling the generation of fine-grained 3D facial motions.

\begin{table}
    \centering
    \caption{\textbf{Experimental results for benchmark II: text-to-3D facial motion generation.} The best and runner-up performances are bold and underlined, respectively.}
    \resizebox{\linewidth}{!}{ 
    \begin{tabular}{l c c c c c c }
    \toprule
    \multirow{2}{*}{Methods}  & \multicolumn{3}{c}{Text R-Precision $\uparrow$} & \multirow{2}{*}{FID $\downarrow$} & \multirow{2}{*}{Text-Match $\downarrow$} & \multirow{2}{*}{Diversity $\rightarrow$} \\

    \cline{2-4}
    ~ & Top-1 & Top-2 & Top-3 \\
    
    \midrule

        Real motion &  0.678  & 0.842 & 0.904 &  0 & 19.14 & 33.39   \\
        VQ-VAE \small{(Recons.)} & 0.492 & 0.667 & 0.763  &  21.29 & 21.72 & 33.48  \\ 
    \midrule

        TM2T \cite{guo2022tm2t}  & 0.237 & 0.369 & 0.460 & 144.0 & 26.23 & 31.01  \\

        T2M-GPT \cite{zhang2023t2m}   & \underline{0.331} &  \underline{0.495} & \underline{0.598} & \textbf{22.36}  & \underline{24.62} &  \underline{32.91}  \\ 

        MDM \cite{tevet2022human}  & 0.214  & 0.327 & 0.404 & 64.45 & 27.82  & 31.82 \\

        \rowcolor[gray]{.92}
        \textbf{Our MM2Face} & \textbf{0.405} & \textbf{0.601} & \textbf{0.698} &  \underline{35.80} & \textbf{23.17} &  \textbf{33.30} \\

    \bottomrule
    \end{tabular} 
    }
    \label{bench2_table}
\end{table}

\subsection{Text-to-3D Facial Motion Generation}

MMHead contains diverse facial expression sequences from specific face expression datasets and general facial motion datasets. Hence we explore a new text-to-3D facial motion generation task. We conduct experiments with the state-of-the-art text-to-3D human motion methods, \textit{i.e.}, TM2T \cite{guo2022tm2t}, T2M-GPT \cite{zhang2023t2m}, and MDM \cite{tevet2022human}. We re-implement these methods to adapt to our dataset format.

\noindent\textbf{Experiment Setup and Metrics.} Similar to benchmark I, we select 56 dimensions FLAME parameters as motion representations and adopt the Frechet Inception Distance (FID), R-precision, Diversity, and Text Matching Score for evaluation. We retrain a text-to-3D facial motion retrieval network in the benchmark II subset for metric calculation, following TMR \cite{petrovich2023tmr}.

\noindent\textbf{Quantitative Evaluation.} The quantitative results are summarized in Tab. \ref{bench2_table} on the benchmark II subset of the MMHead dataset. We can derive that our method achieves the best performance against all metrics. T2M-GPT \cite{zhang2023t2m} achieves the second-best performance in FID, matching score, diversity, and R-precision. MDM \cite{tevet2022human} achieves considerable FID performance but fails to capture the text-to-motion consistency according to R-precision. From the results, we derive that our MMHead dataset has the potential for further explorations.

\noindent\textbf{Qualitative Evaluation.} We also provide the qualitative results in Fig. \ref{fig:t2m_results}. Our approach exhibits superior text and face alignment. More results are in the supplementary material.

\begin{figure}
\centering
\includegraphics[width=\linewidth]{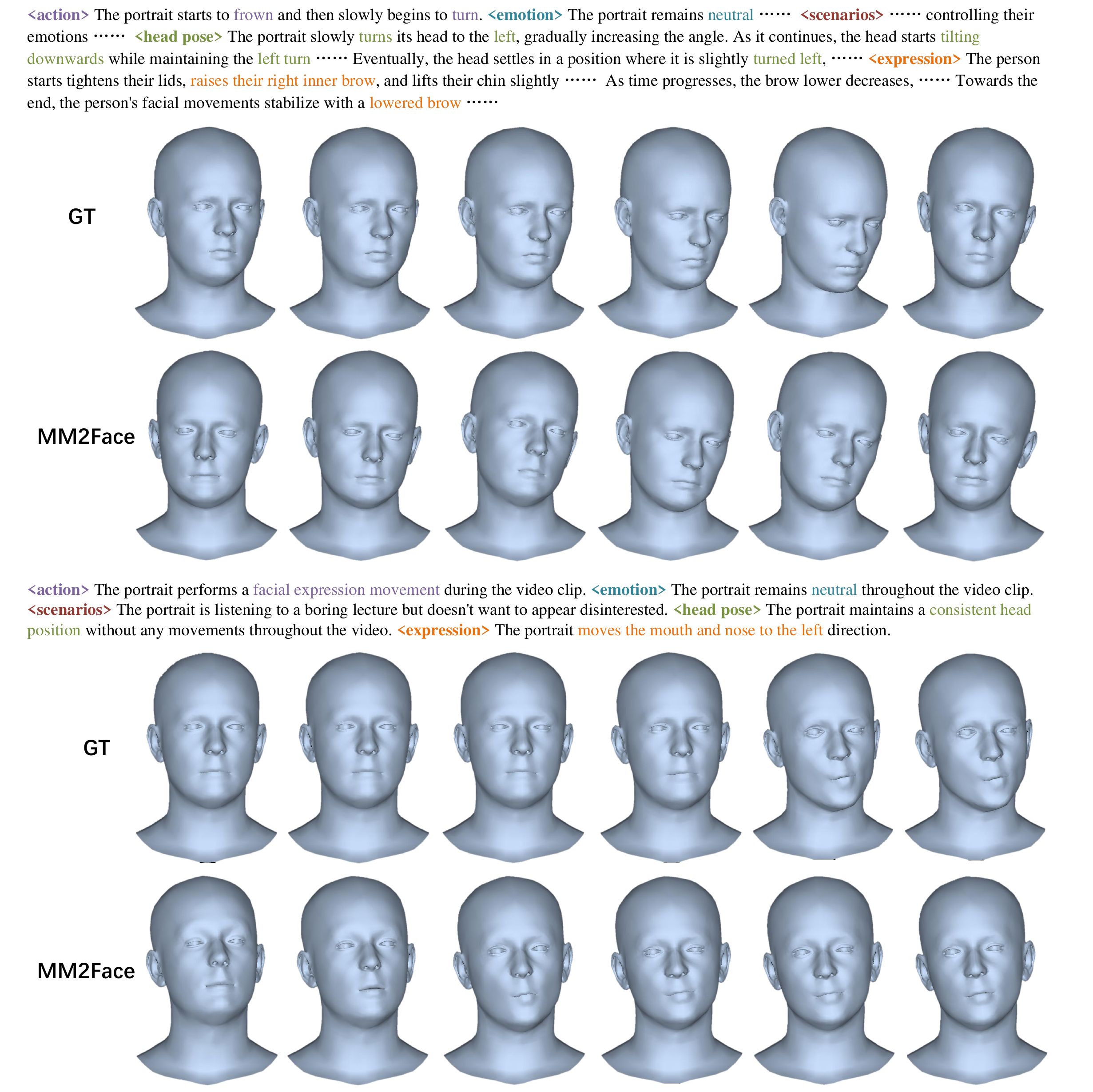}
\vspace{-1mm}
\caption{Qualitative results of MM2Face on benchmark II: text-to-3D facial motion generation.}
\label{fig:t2m_results}
\vspace{-4mm}
\end{figure}

\section{Conclusion}
In this paper, we push forward the 3D facial animation task, and present MMHead, the first multi-modal 3D facial animation dataset with rich hierarchical text annotations including abstract action and emotion descriptions, fine-grained facial and head movements descriptions, and possible emotion scenarios.
With MMHead, we benchmark two new tasks: text-induced 3D talking head animation and text-to-3D facial motion generation. Moreover, we propose a simple but efficient VQ-VAE-based method named MM2Face to explore the multi-modal information fusion strategies and generate diverse and plausible 3D facial motions, which achieves competitive performance on both benchmarks. 
We hope that the MMHead dataset and the corresponding benchmarks will promote in-depth research works on multi-modal 3D facial animation.

\begin{acks}
This work was supported by NSFC 62225112 and Shanghai Municipal Science and Technology Major Project (2021SHZDZX0102).
\end{acks}

\bibliographystyle{ACM-Reference-Format}
\balance
\bibliography{ref}


\begin{thebibliography}{88}


\ifx \showCODEN    \undefined \def \showCODEN     #1{\unskip}     \fi
\ifx \showDOI      \undefined \def \showDOI       #1{#1}\fi
\ifx \showISBNx    \undefined \def \showISBNx     #1{\unskip}     \fi
\ifx \showISBNxiii \undefined \def \showISBNxiii  #1{\unskip}     \fi
\ifx \showISSN     \undefined \def \showISSN      #1{\unskip}     \fi
\ifx \showLCCN     \undefined \def \showLCCN      #1{\unskip}     \fi
\ifx \shownote     \undefined \def \shownote      #1{#1}          \fi
\ifx \showarticletitle \undefined \def \showarticletitle #1{#1}   \fi
\ifx \showURL      \undefined \def \showURL       {\relax}        \fi
\providecommand\bibfield[2]{#2}
\providecommand\bibinfo[2]{#2}
\providecommand\natexlab[1]{#1}
\providecommand\showeprint[2][]{arXiv:#2}

\bibitem[Aneja et~al\mbox{.}(2024)]%
        {aneja2024facetalk}
\bibfield{author}{\bibinfo{person}{Shivangi Aneja}, \bibinfo{person}{Justus Thies}, \bibinfo{person}{Angela Dai}, {and} \bibinfo{person}{Matthias Nie{\ss}ner}.} \bibinfo{year}{2024}\natexlab{}.
\newblock \showarticletitle{Facetalk: Audio-driven motion diffusion for neural parametric head models}. In \bibinfo{booktitle}{\emph{Proceedings of the IEEE/CVF Conference on Computer Vision and Pattern Recognition}}. \bibinfo{pages}{21263--21273}.
\newblock


\bibitem[Baevski et~al\mbox{.}(2020)]%
        {baevski2020wav2vec}
\bibfield{author}{\bibinfo{person}{Alexei Baevski}, \bibinfo{person}{Yuhao Zhou}, \bibinfo{person}{Abdelrahman Mohamed}, {and} \bibinfo{person}{Michael Auli}.} \bibinfo{year}{2020}\natexlab{}.
\newblock \showarticletitle{wav2vec 2.0: A framework for self-supervised learning of speech representations}.
\newblock \bibinfo{journal}{\emph{Advances in neural information processing systems}}  \bibinfo{volume}{33} (\bibinfo{year}{2020}), \bibinfo{pages}{12449--12460}.
\newblock


\bibitem[Betker et~al\mbox{.}(2023)]%
        {betker2023improving}
\bibfield{author}{\bibinfo{person}{James Betker}, \bibinfo{person}{Gabriel Goh}, \bibinfo{person}{Li Jing}, \bibinfo{person}{Tim Brooks}, \bibinfo{person}{Jianfeng Wang}, \bibinfo{person}{Linjie Li}, \bibinfo{person}{Long Ouyang}, \bibinfo{person}{Juntang Zhuang}, \bibinfo{person}{Joyce Lee}, \bibinfo{person}{Yufei Guo}, {et~al\mbox{.}}} \bibinfo{year}{2023}\natexlab{}.
\newblock \showarticletitle{Improving image generation with better captions}.
\newblock \bibinfo{journal}{\emph{Computer Science. https://cdn. openai. com/papers/dall-e-3. pdf}} \bibinfo{volume}{2}, \bibinfo{number}{3} (\bibinfo{year}{2023}), \bibinfo{pages}{8}.
\newblock


\bibitem[Chang et~al\mbox{.}(2023)]%
        {chang2023muse}
\bibfield{author}{\bibinfo{person}{Huiwen Chang}, \bibinfo{person}{Han Zhang}, \bibinfo{person}{Jarred Barber}, \bibinfo{person}{AJ Maschinot}, \bibinfo{person}{Jose Lezama}, \bibinfo{person}{Lu Jiang}, \bibinfo{person}{Ming-Hsuan Yang}, \bibinfo{person}{Kevin Murphy}, \bibinfo{person}{William~T Freeman}, \bibinfo{person}{Michael Rubinstein}, {et~al\mbox{.}}} \bibinfo{year}{2023}\natexlab{}.
\newblock \showarticletitle{Muse: Text-to-image generation via masked generative transformers}.
\newblock \bibinfo{journal}{\emph{arXiv preprint arXiv:2301.00704}} (\bibinfo{year}{2023}).
\newblock


\bibitem[Chen et~al\mbox{.}(2023)]%
        {chen2023executing}
\bibfield{author}{\bibinfo{person}{Xin Chen}, \bibinfo{person}{Biao Jiang}, \bibinfo{person}{Wen Liu}, \bibinfo{person}{Zilong Huang}, \bibinfo{person}{Bin Fu}, \bibinfo{person}{Tao Chen}, {and} \bibinfo{person}{Gang Yu}.} \bibinfo{year}{2023}\natexlab{}.
\newblock \showarticletitle{Executing your commands via motion diffusion in latent space}. In \bibinfo{booktitle}{\emph{Proceedings of the IEEE/CVF Conference on Computer Vision and Pattern Recognition}}. \bibinfo{pages}{18000--18010}.
\newblock


\bibitem[Cheng et~al\mbox{.}(2018)]%
        {cheng20184dfab}
\bibfield{author}{\bibinfo{person}{Shiyang Cheng}, \bibinfo{person}{Irene Kotsia}, \bibinfo{person}{Maja Pantic}, {and} \bibinfo{person}{Stefanos Zafeiriou}.} \bibinfo{year}{2018}\natexlab{}.
\newblock \showarticletitle{4dfab: A large scale 4d database for facial expression analysis and biometric applications}. In \bibinfo{booktitle}{\emph{Proceedings of the IEEE conference on computer vision and pattern recognition}}. \bibinfo{pages}{5117--5126}.
\newblock


\bibitem[Chung et~al\mbox{.}(2018)]%
        {chung2018voxceleb2}
\bibfield{author}{\bibinfo{person}{Joon~Son Chung}, \bibinfo{person}{Arsha Nagrani}, {and} \bibinfo{person}{Andrew Zisserman}.} \bibinfo{year}{2018}\natexlab{}.
\newblock \showarticletitle{Voxceleb2: Deep speaker recognition}.
\newblock \bibinfo{journal}{\emph{arXiv preprint arXiv:1806.05622}} (\bibinfo{year}{2018}).
\newblock


\bibitem[Chung and Zisserman(2017)]%
        {chung2017out}
\bibfield{author}{\bibinfo{person}{Joon~Son Chung} {and} \bibinfo{person}{Andrew Zisserman}.} \bibinfo{year}{2017}\natexlab{}.
\newblock \showarticletitle{Out of time: automated lip sync in the wild}. In \bibinfo{booktitle}{\emph{Computer Vision--ACCV 2016 Workshops: ACCV 2016 International Workshops, Taipei, Taiwan, November 20-24, 2016, Revised Selected Papers, Part II 13}}. Springer, \bibinfo{pages}{251--263}.
\newblock


\bibitem[Cohen et~al\mbox{.}(2001)]%
        {cohen2001animated}
\bibfield{author}{\bibinfo{person}{Michael~M Cohen}, \bibinfo{person}{Rashid Clark}, {and} \bibinfo{person}{Dominic~W Massaro}.} \bibinfo{year}{2001}\natexlab{}.
\newblock \showarticletitle{Animated speech: Research progress and applications}. In \bibinfo{booktitle}{\emph{AVSP 2001-International Conference on Auditory-Visual Speech Processing}}.
\newblock


\bibitem[Cosker et~al\mbox{.}(2011)]%
        {cosker2011facs}
\bibfield{author}{\bibinfo{person}{Darren Cosker}, \bibinfo{person}{Eva Krumhuber}, {and} \bibinfo{person}{Adrian Hilton}.} \bibinfo{year}{2011}\natexlab{}.
\newblock \showarticletitle{A FACS valid 3D dynamic action unit database with applications to 3D dynamic morphable facial modeling}. In \bibinfo{booktitle}{\emph{2011 international conference on computer vision}}. IEEE, \bibinfo{pages}{2296--2303}.
\newblock


\bibitem[Cudeiro et~al\mbox{.}(2019)]%
        {cudeiro2019capture}
\bibfield{author}{\bibinfo{person}{Daniel Cudeiro}, \bibinfo{person}{Timo Bolkart}, \bibinfo{person}{Cassidy Laidlaw}, \bibinfo{person}{Anurag Ranjan}, {and} \bibinfo{person}{Michael~J Black}.} \bibinfo{year}{2019}\natexlab{}.
\newblock \showarticletitle{Capture, learning, and synthesis of 3D speaking styles}. In \bibinfo{booktitle}{\emph{Proceedings of the IEEE/CVF Conference on Computer Vision and Pattern Recognition}}. \bibinfo{pages}{10101--10111}.
\newblock


\bibitem[Dan{\v{e}}{\v{c}}ek et~al\mbox{.}(2022)]%
        {danvevcek2022emoca}
\bibfield{author}{\bibinfo{person}{Radek Dan{\v{e}}{\v{c}}ek}, \bibinfo{person}{Michael~J Black}, {and} \bibinfo{person}{Timo Bolkart}.} \bibinfo{year}{2022}\natexlab{}.
\newblock \showarticletitle{EMOCA: Emotion driven monocular face capture and animation}. In \bibinfo{booktitle}{\emph{Proceedings of the IEEE/CVF Conference on Computer Vision and Pattern Recognition}}. \bibinfo{pages}{20311--20322}.
\newblock


\bibitem[Dan{\v{e}}{\v{c}}ek et~al\mbox{.}(2023)]%
        {danvevcek2023emotional}
\bibfield{author}{\bibinfo{person}{Radek Dan{\v{e}}{\v{c}}ek}, \bibinfo{person}{Kiran Chhatre}, \bibinfo{person}{Shashank Tripathi}, \bibinfo{person}{Yandong Wen}, \bibinfo{person}{Michael Black}, {and} \bibinfo{person}{Timo Bolkart}.} \bibinfo{year}{2023}\natexlab{}.
\newblock \showarticletitle{Emotional speech-driven animation with content-emotion disentanglement}. In \bibinfo{booktitle}{\emph{SIGGRAPH Asia 2023 Conference Papers}}. \bibinfo{pages}{1--13}.
\newblock


\bibitem[Edwards et~al\mbox{.}(2016)]%
        {edwards2016jali}
\bibfield{author}{\bibinfo{person}{Pif Edwards}, \bibinfo{person}{Chris Landreth}, \bibinfo{person}{Eugene Fiume}, {and} \bibinfo{person}{Karan Singh}.} \bibinfo{year}{2016}\natexlab{}.
\newblock \showarticletitle{Jali: an animator-centric viseme model for expressive lip synchronization}.
\newblock \bibinfo{journal}{\emph{ACM Transactions on graphics (TOG)}} \bibinfo{volume}{35}, \bibinfo{number}{4} (\bibinfo{year}{2016}), \bibinfo{pages}{1--11}.
\newblock


\bibitem[Esser et~al\mbox{.}(2021)]%
        {esser2021taming}
\bibfield{author}{\bibinfo{person}{Patrick Esser}, \bibinfo{person}{Robin Rombach}, {and} \bibinfo{person}{Bjorn Ommer}.} \bibinfo{year}{2021}\natexlab{}.
\newblock \showarticletitle{Taming transformers for high-resolution image synthesis}. In \bibinfo{booktitle}{\emph{Proceedings of the IEEE/CVF conference on computer vision and pattern recognition}}. \bibinfo{pages}{12873--12883}.
\newblock


\bibitem[Fan et~al\mbox{.}(2022)]%
        {fan2022faceformer}
\bibfield{author}{\bibinfo{person}{Yingruo Fan}, \bibinfo{person}{Zhaojiang Lin}, \bibinfo{person}{Jun Saito}, \bibinfo{person}{Wenping Wang}, {and} \bibinfo{person}{Taku Komura}.} \bibinfo{year}{2022}\natexlab{}.
\newblock \showarticletitle{Faceformer: Speech-driven 3d facial animation with transformers}. In \bibinfo{booktitle}{\emph{Proceedings of the IEEE/CVF Conference on Computer Vision and Pattern Recognition}}. \bibinfo{pages}{18770--18780}.
\newblock


\bibitem[Fanelli et~al\mbox{.}(2010)]%
        {fanelli20103}
\bibfield{author}{\bibinfo{person}{Gabriele Fanelli}, \bibinfo{person}{Juergen Gall}, \bibinfo{person}{Harald Romsdorfer}, \bibinfo{person}{Thibaut Weise}, {and} \bibinfo{person}{Luc Van~Gool}.} \bibinfo{year}{2010}\natexlab{}.
\newblock \showarticletitle{A 3-d audio-visual corpus of affective communication}.
\newblock \bibinfo{journal}{\emph{IEEE Transactions on Multimedia}} \bibinfo{volume}{12}, \bibinfo{number}{6} (\bibinfo{year}{2010}), \bibinfo{pages}{591--598}.
\newblock


\bibitem[Feng et~al\mbox{.}(2021)]%
        {DECA:Siggraph2021}
\bibfield{author}{\bibinfo{person}{Yao Feng}, \bibinfo{person}{Haiwen Feng}, \bibinfo{person}{Michael~J. Black}, {and} \bibinfo{person}{Timo Bolkart}.} \bibinfo{year}{2021}\natexlab{}.
\newblock \showarticletitle{Learning an Animatable Detailed {3D} Face Model from In-The-Wild Images}.
\newblock \bibinfo{journal}{\emph{ACM Transactions on Graphics (ToG), Proc. SIGGRAPH}} \bibinfo{volume}{40}, \bibinfo{number}{8} (\bibinfo{year}{2021}).
\newblock
\urldef\tempurl%
\url{https://doi.org/10.1145/3450626.3459936}
\showURL{%
\tempurl}


\bibitem[Filntisis et~al\mbox{.}(2022)]%
        {filntisis2022visual}
\bibfield{author}{\bibinfo{person}{Panagiotis~P. Filntisis}, \bibinfo{person}{George Retsinas}, \bibinfo{person}{Foivos Paraperas-Papantoniou}, \bibinfo{person}{Athanasios Katsamanis}, \bibinfo{person}{Anastasios Roussos}, {and} \bibinfo{person}{Petros Maragos}.} \bibinfo{year}{2022}\natexlab{}.
\newblock \showarticletitle{Visual Speech-Aware Perceptual 3D Facial Expression Reconstruction from Videos}.
\newblock \bibinfo{journal}{\emph{arXiv preprint arXiv:2207.11094}} (\bibinfo{year}{2022}).
\newblock


\bibitem[Goodfellow et~al\mbox{.}(2020)]%
        {goodfellow2020generative}
\bibfield{author}{\bibinfo{person}{Ian Goodfellow}, \bibinfo{person}{Jean Pouget-Abadie}, \bibinfo{person}{Mehdi Mirza}, \bibinfo{person}{Bing Xu}, \bibinfo{person}{David Warde-Farley}, \bibinfo{person}{Sherjil Ozair}, \bibinfo{person}{Aaron Courville}, {and} \bibinfo{person}{Yoshua Bengio}.} \bibinfo{year}{2020}\natexlab{}.
\newblock \showarticletitle{Generative adversarial networks}.
\newblock \bibinfo{journal}{\emph{Commun. ACM}} \bibinfo{volume}{63}, \bibinfo{number}{11} (\bibinfo{year}{2020}), \bibinfo{pages}{139--144}.
\newblock


\bibitem[Guo et~al\mbox{.}(2023)]%
        {guo2023momask}
\bibfield{author}{\bibinfo{person}{Chuan Guo}, \bibinfo{person}{Yuxuan Mu}, \bibinfo{person}{Muhammad~Gohar Javed}, \bibinfo{person}{Sen Wang}, {and} \bibinfo{person}{Li Cheng}.} \bibinfo{year}{2023}\natexlab{}.
\newblock \showarticletitle{MoMask: Generative Masked Modeling of 3D Human Motions}.
\newblock \bibinfo{journal}{\emph{arXiv preprint arXiv:2312.00063}} (\bibinfo{year}{2023}).
\newblock


\bibitem[Guo et~al\mbox{.}(2022a)]%
        {guo2022generating}
\bibfield{author}{\bibinfo{person}{Chuan Guo}, \bibinfo{person}{Shihao Zou}, \bibinfo{person}{Xinxin Zuo}, \bibinfo{person}{Sen Wang}, \bibinfo{person}{Wei Ji}, \bibinfo{person}{Xingyu Li}, {and} \bibinfo{person}{Li Cheng}.} \bibinfo{year}{2022}\natexlab{a}.
\newblock \showarticletitle{Generating diverse and natural 3d human motions from text}. In \bibinfo{booktitle}{\emph{Proceedings of the IEEE/CVF Conference on Computer Vision and Pattern Recognition}}. \bibinfo{pages}{5152--5161}.
\newblock


\bibitem[Guo et~al\mbox{.}(2022b)]%
        {guo2022tm2t}
\bibfield{author}{\bibinfo{person}{Chuan Guo}, \bibinfo{person}{Xinxin Zuo}, \bibinfo{person}{Sen Wang}, {and} \bibinfo{person}{Li Cheng}.} \bibinfo{year}{2022}\natexlab{b}.
\newblock \showarticletitle{Tm2t: Stochastic and tokenized modeling for the reciprocal generation of 3d human motions and texts}. In \bibinfo{booktitle}{\emph{European Conference on Computer Vision}}. Springer, \bibinfo{pages}{580--597}.
\newblock


\bibitem[Guo et~al\mbox{.}(2020)]%
        {guo2020action2motion}
\bibfield{author}{\bibinfo{person}{Chuan Guo}, \bibinfo{person}{Xinxin Zuo}, \bibinfo{person}{Sen Wang}, \bibinfo{person}{Shihao Zou}, \bibinfo{person}{Qingyao Sun}, \bibinfo{person}{Annan Deng}, \bibinfo{person}{Minglun Gong}, {and} \bibinfo{person}{Li Cheng}.} \bibinfo{year}{2020}\natexlab{}.
\newblock \showarticletitle{Action2motion: Conditioned generation of 3d human motions}. In \bibinfo{booktitle}{\emph{Proceedings of the 28th ACM International Conference on Multimedia}}. \bibinfo{pages}{2021--2029}.
\newblock


\bibitem[He et~al\mbox{.}(2016)]%
        {he2016deep}
\bibfield{author}{\bibinfo{person}{Kaiming He}, \bibinfo{person}{Xiangyu Zhang}, \bibinfo{person}{Shaoqing Ren}, {and} \bibinfo{person}{Jian Sun}.} \bibinfo{year}{2016}\natexlab{}.
\newblock \showarticletitle{Deep residual learning for image recognition}. In \bibinfo{booktitle}{\emph{Proceedings of the IEEE conference on computer vision and pattern recognition}}. \bibinfo{pages}{770--778}.
\newblock


\bibitem[He et~al\mbox{.}(2023)]%
        {he2023speech4mesh}
\bibfield{author}{\bibinfo{person}{Shan He}, \bibinfo{person}{Haonan He}, \bibinfo{person}{Shuo Yang}, \bibinfo{person}{Xiaoyan Wu}, \bibinfo{person}{Pengcheng Xia}, \bibinfo{person}{Bing Yin}, \bibinfo{person}{Cong Liu}, \bibinfo{person}{Lirong Dai}, {and} \bibinfo{person}{Chang Xu}.} \bibinfo{year}{2023}\natexlab{}.
\newblock \showarticletitle{Speech4Mesh: Speech-Assisted Monocular 3D Facial Reconstruction for Speech-Driven 3D Facial Animation}. In \bibinfo{booktitle}{\emph{Proceedings of the IEEE/CVF International Conference on Computer Vision}}. \bibinfo{pages}{14192--14202}.
\newblock


\bibitem[Ho et~al\mbox{.}(2020)]%
        {ho2020denoising}
\bibfield{author}{\bibinfo{person}{Jonathan Ho}, \bibinfo{person}{Ajay Jain}, {and} \bibinfo{person}{Pieter Abbeel}.} \bibinfo{year}{2020}\natexlab{}.
\newblock \showarticletitle{Denoising diffusion probabilistic models}.
\newblock \bibinfo{journal}{\emph{Advances in neural information processing systems}}  \bibinfo{volume}{33} (\bibinfo{year}{2020}), \bibinfo{pages}{6840--6851}.
\newblock


\bibitem[Jiang et~al\mbox{.}(2024)]%
        {jiang2024motiongpt}
\bibfield{author}{\bibinfo{person}{Biao Jiang}, \bibinfo{person}{Xin Chen}, \bibinfo{person}{Wen Liu}, \bibinfo{person}{Jingyi Yu}, \bibinfo{person}{Gang Yu}, {and} \bibinfo{person}{Tao Chen}.} \bibinfo{year}{2024}\natexlab{}.
\newblock \showarticletitle{Motiongpt: Human motion as a foreign language}.
\newblock \bibinfo{journal}{\emph{Advances in Neural Information Processing Systems}}  \bibinfo{volume}{36} (\bibinfo{year}{2024}).
\newblock


\bibitem[Kang et~al\mbox{.}(2023)]%
        {kang2023scaling}
\bibfield{author}{\bibinfo{person}{Minguk Kang}, \bibinfo{person}{Jun-Yan Zhu}, \bibinfo{person}{Richard Zhang}, \bibinfo{person}{Jaesik Park}, \bibinfo{person}{Eli Shechtman}, \bibinfo{person}{Sylvain Paris}, {and} \bibinfo{person}{Taesung Park}.} \bibinfo{year}{2023}\natexlab{}.
\newblock \showarticletitle{Scaling up gans for text-to-image synthesis}. In \bibinfo{booktitle}{\emph{Proceedings of the IEEE/CVF Conference on Computer Vision and Pattern Recognition}}. \bibinfo{pages}{10124--10134}.
\newblock


\bibitem[Kingma and Ba(2014)]%
        {kingma2014adam}
\bibfield{author}{\bibinfo{person}{Diederik~P Kingma} {and} \bibinfo{person}{Jimmy Ba}.} \bibinfo{year}{2014}\natexlab{}.
\newblock \showarticletitle{Adam: A method for stochastic optimization}.
\newblock \bibinfo{journal}{\emph{arXiv preprint arXiv:1412.6980}} (\bibinfo{year}{2014}).
\newblock


\bibitem[Kingma and Welling(2013)]%
        {kingma2013auto}
\bibfield{author}{\bibinfo{person}{Diederik~P Kingma} {and} \bibinfo{person}{Max Welling}.} \bibinfo{year}{2013}\natexlab{}.
\newblock \showarticletitle{Auto-encoding variational bayes}.
\newblock \bibinfo{journal}{\emph{arXiv preprint arXiv:1312.6114}} (\bibinfo{year}{2013}).
\newblock


\bibitem[Li et~al\mbox{.}(2017)]%
        {li2017learning}
\bibfield{author}{\bibinfo{person}{Tianye Li}, \bibinfo{person}{Timo Bolkart}, \bibinfo{person}{Michael~J Black}, \bibinfo{person}{Hao Li}, {and} \bibinfo{person}{Javier Romero}.} \bibinfo{year}{2017}\natexlab{}.
\newblock \showarticletitle{Learning a model of facial shape and expression from 4D scans.}
\newblock \bibinfo{journal}{\emph{ACM Trans. Graph.}} \bibinfo{volume}{36}, \bibinfo{number}{6} (\bibinfo{year}{2017}), \bibinfo{pages}{194--1}.
\newblock


\bibitem[Liang et~al\mbox{.}(2024)]%
        {liang2024intergen}
\bibfield{author}{\bibinfo{person}{Han Liang}, \bibinfo{person}{Wenqian Zhang}, \bibinfo{person}{Wenxuan Li}, \bibinfo{person}{Jingyi Yu}, {and} \bibinfo{person}{Lan Xu}.} \bibinfo{year}{2024}\natexlab{}.
\newblock \showarticletitle{Intergen: Diffusion-based multi-human motion generation under complex interactions}.
\newblock \bibinfo{journal}{\emph{International Journal of Computer Vision}} (\bibinfo{year}{2024}), \bibinfo{pages}{1--21}.
\newblock


\bibitem[Lin et~al\mbox{.}(2024)]%
        {lin2024motion}
\bibfield{author}{\bibinfo{person}{Jing Lin}, \bibinfo{person}{Ailing Zeng}, \bibinfo{person}{Shunlin Lu}, \bibinfo{person}{Yuanhao Cai}, \bibinfo{person}{Ruimao Zhang}, \bibinfo{person}{Haoqian Wang}, {and} \bibinfo{person}{Lei Zhang}.} \bibinfo{year}{2024}\natexlab{}.
\newblock \showarticletitle{Motion-x: A large-scale 3d expressive whole-body human motion dataset}.
\newblock \bibinfo{journal}{\emph{Advances in Neural Information Processing Systems}}  \bibinfo{volume}{36} (\bibinfo{year}{2024}).
\newblock


\bibitem[Livingstone and Russo(2018)]%
        {livingstone2018ryerson}
\bibfield{author}{\bibinfo{person}{Steven~R Livingstone} {and} \bibinfo{person}{Frank~A Russo}.} \bibinfo{year}{2018}\natexlab{}.
\newblock \showarticletitle{The Ryerson Audio-Visual Database of Emotional Speech and Song (RAVDESS): A dynamic, multimodal set of facial and vocal expressions in North American English}.
\newblock \bibinfo{journal}{\emph{PloS one}} \bibinfo{volume}{13}, \bibinfo{number}{5} (\bibinfo{year}{2018}), \bibinfo{pages}{e0196391}.
\newblock


\bibitem[Luo et~al\mbox{.}(2022)]%
        {luo2022learning}
\bibfield{author}{\bibinfo{person}{Cheng Luo}, \bibinfo{person}{Siyang Song}, \bibinfo{person}{Weicheng Xie}, \bibinfo{person}{Linlin Shen}, {and} \bibinfo{person}{Hatice Gunes}.} \bibinfo{year}{2022}\natexlab{}.
\newblock \showarticletitle{Learning multi-dimensional edge feature-based au relation graph for facial action unit recognition}.
\newblock \bibinfo{journal}{\emph{arXiv preprint arXiv:2205.01782}} (\bibinfo{year}{2022}).
\newblock


\bibitem[Ma et~al\mbox{.}(2023)]%
        {ma2023talkclip}
\bibfield{author}{\bibinfo{person}{Yifeng Ma}, \bibinfo{person}{Suzhen Wang}, \bibinfo{person}{Yu Ding}, \bibinfo{person}{Bowen Ma}, \bibinfo{person}{Tangjie Lv}, \bibinfo{person}{Changjie Fan}, \bibinfo{person}{Zhipeng Hu}, \bibinfo{person}{Zhidong Deng}, {and} \bibinfo{person}{Xin Yu}.} \bibinfo{year}{2023}\natexlab{}.
\newblock \showarticletitle{TalkCLIP: Talking Head Generation with Text-Guided Expressive Speaking Styles}.
\newblock \bibinfo{journal}{\emph{arXiv preprint arXiv:2304.00334}} (\bibinfo{year}{2023}).
\newblock


\bibitem[OpenAI(2023)]%
        {chatgpt}
\bibfield{author}{\bibinfo{person}{OpenAI}.} \bibinfo{year}{2023}\natexlab{}.
\newblock \showarticletitle{ChatGPT. https://chat.openai.com/chat. Accessed: 2023-05-14}.
\newblock  (\bibinfo{year}{2023}).
\newblock


\bibitem[Otberdout et~al\mbox{.}(2022)]%
        {otberdout2022sparse}
\bibfield{author}{\bibinfo{person}{Naima Otberdout}, \bibinfo{person}{Claudio Ferrari}, \bibinfo{person}{Mohamed Daoudi}, \bibinfo{person}{Stefano Berretti}, {and} \bibinfo{person}{Alberto Del~Bimbo}.} \bibinfo{year}{2022}\natexlab{}.
\newblock \showarticletitle{Sparse to dense dynamic 3d facial expression generation}. In \bibinfo{booktitle}{\emph{Proceedings of the IEEE/CVF conference on computer vision and pattern recognition}}. \bibinfo{pages}{20385--20394}.
\newblock


\bibitem[Pan et~al\mbox{.}(2024)]%
        {pan2024renderme}
\bibfield{author}{\bibinfo{person}{Dongwei Pan}, \bibinfo{person}{Long Zhuo}, \bibinfo{person}{Jingtan Piao}, \bibinfo{person}{Huiwen Luo}, \bibinfo{person}{Wei Cheng}, \bibinfo{person}{Yuxin Wang}, \bibinfo{person}{Siming Fan}, \bibinfo{person}{Shengqi Liu}, \bibinfo{person}{Lei Yang}, \bibinfo{person}{Bo Dai}, {et~al\mbox{.}}} \bibinfo{year}{2024}\natexlab{}.
\newblock \showarticletitle{RenderMe-360: A Large Digital Asset Library and Benchmarks Towards High-fidelity Head Avatars}.
\newblock \bibinfo{journal}{\emph{Advances in Neural Information Processing Systems}}  \bibinfo{volume}{36} (\bibinfo{year}{2024}).
\newblock


\bibitem[Peng et~al\mbox{.}(2023a)]%
        {peng2023selftalk}
\bibfield{author}{\bibinfo{person}{Ziqiao Peng}, \bibinfo{person}{Yihao Luo}, \bibinfo{person}{Yue Shi}, \bibinfo{person}{Hao Xu}, \bibinfo{person}{Xiangyu Zhu}, \bibinfo{person}{Hongyan Liu}, \bibinfo{person}{Jun He}, {and} \bibinfo{person}{Zhaoxin Fan}.} \bibinfo{year}{2023}\natexlab{a}.
\newblock \showarticletitle{Selftalk: A self-supervised commutative training diagram to comprehend 3d talking faces}. In \bibinfo{booktitle}{\emph{Proceedings of the 31st ACM International Conference on Multimedia}}. \bibinfo{pages}{5292--5301}.
\newblock


\bibitem[Peng et~al\mbox{.}(2023b)]%
        {peng2023emotalk}
\bibfield{author}{\bibinfo{person}{Ziqiao Peng}, \bibinfo{person}{Haoyu Wu}, \bibinfo{person}{Zhenbo Song}, \bibinfo{person}{Hao Xu}, \bibinfo{person}{Xiangyu Zhu}, \bibinfo{person}{Jun He}, \bibinfo{person}{Hongyan Liu}, {and} \bibinfo{person}{Zhaoxin Fan}.} \bibinfo{year}{2023}\natexlab{b}.
\newblock \showarticletitle{Emotalk: Speech-driven emotional disentanglement for 3d face animation}. In \bibinfo{booktitle}{\emph{Proceedings of the IEEE/CVF International Conference on Computer Vision}}. \bibinfo{pages}{20687--20697}.
\newblock


\bibitem[Petrovich et~al\mbox{.}(2023)]%
        {petrovich2023tmr}
\bibfield{author}{\bibinfo{person}{Mathis Petrovich}, \bibinfo{person}{Michael~J Black}, {and} \bibinfo{person}{G{\"u}l Varol}.} \bibinfo{year}{2023}\natexlab{}.
\newblock \showarticletitle{TMR: Text-to-motion retrieval using contrastive 3D human motion synthesis}. In \bibinfo{booktitle}{\emph{Proceedings of the IEEE/CVF International Conference on Computer Vision}}. \bibinfo{pages}{9488--9497}.
\newblock


\bibitem[Pinyoanuntapong et~al\mbox{.}(2023)]%
        {pinyoanuntapong2023mmm}
\bibfield{author}{\bibinfo{person}{Ekkasit Pinyoanuntapong}, \bibinfo{person}{Pu Wang}, \bibinfo{person}{Minwoo Lee}, {and} \bibinfo{person}{Chen Chen}.} \bibinfo{year}{2023}\natexlab{}.
\newblock \showarticletitle{Mmm: Generative masked motion model}.
\newblock \bibinfo{journal}{\emph{arXiv preprint arXiv:2312.03596}} (\bibinfo{year}{2023}).
\newblock


\bibitem[Ranjan et~al\mbox{.}(2018)]%
        {ranjan2018generating}
\bibfield{author}{\bibinfo{person}{Anurag Ranjan}, \bibinfo{person}{Timo Bolkart}, \bibinfo{person}{Soubhik Sanyal}, {and} \bibinfo{person}{Michael~J Black}.} \bibinfo{year}{2018}\natexlab{}.
\newblock \showarticletitle{Generating 3D faces using convolutional mesh autoencoders}. In \bibinfo{booktitle}{\emph{Proceedings of the European conference on computer vision (ECCV)}}. \bibinfo{pages}{704--720}.
\newblock


\bibitem[Richard et~al\mbox{.}(2021)]%
        {richard2021meshtalk}
\bibfield{author}{\bibinfo{person}{Alexander Richard}, \bibinfo{person}{Michael Zollh{\"o}fer}, \bibinfo{person}{Yandong Wen}, \bibinfo{person}{Fernando De~la Torre}, {and} \bibinfo{person}{Yaser Sheikh}.} \bibinfo{year}{2021}\natexlab{}.
\newblock \showarticletitle{Meshtalk: 3d face animation from speech using cross-modality disentanglement}. In \bibinfo{booktitle}{\emph{Proceedings of the IEEE/CVF International Conference on Computer Vision}}. \bibinfo{pages}{1173--1182}.
\newblock


\bibitem[Rombach et~al\mbox{.}(2022)]%
        {rombach2022high}
\bibfield{author}{\bibinfo{person}{Robin Rombach}, \bibinfo{person}{Andreas Blattmann}, \bibinfo{person}{Dominik Lorenz}, \bibinfo{person}{Patrick Esser}, {and} \bibinfo{person}{Bj{\"o}rn Ommer}.} \bibinfo{year}{2022}\natexlab{}.
\newblock \showarticletitle{High-resolution image synthesis with latent diffusion models}. In \bibinfo{booktitle}{\emph{Proceedings of the IEEE/CVF conference on computer vision and pattern recognition}}. \bibinfo{pages}{10684--10695}.
\newblock


\bibitem[Sanh et~al\mbox{.}(2019)]%
        {sanh2019distilbert}
\bibfield{author}{\bibinfo{person}{Victor Sanh}, \bibinfo{person}{Lysandre Debut}, \bibinfo{person}{Julien Chaumond}, {and} \bibinfo{person}{Thomas Wolf}.} \bibinfo{year}{2019}\natexlab{}.
\newblock \showarticletitle{DistilBERT, a distilled version of BERT: smaller, faster, cheaper and lighter}.
\newblock \bibinfo{journal}{\emph{arXiv preprint arXiv:1910.01108}} (\bibinfo{year}{2019}).
\newblock


\bibitem[Singer et~al\mbox{.}(2022)]%
        {singer2022make}
\bibfield{author}{\bibinfo{person}{Uriel Singer}, \bibinfo{person}{Adam Polyak}, \bibinfo{person}{Thomas Hayes}, \bibinfo{person}{Xi Yin}, \bibinfo{person}{Jie An}, \bibinfo{person}{Songyang Zhang}, \bibinfo{person}{Qiyuan Hu}, \bibinfo{person}{Harry Yang}, \bibinfo{person}{Oron Ashual}, \bibinfo{person}{Oran Gafni}, {et~al\mbox{.}}} \bibinfo{year}{2022}\natexlab{}.
\newblock \showarticletitle{Make-a-video: Text-to-video generation without text-video data}.
\newblock \bibinfo{journal}{\emph{arXiv preprint arXiv:2209.14792}} (\bibinfo{year}{2022}).
\newblock


\bibitem[Stan et~al\mbox{.}(2023)]%
        {stan2023facediffuser}
\bibfield{author}{\bibinfo{person}{Stefan Stan}, \bibinfo{person}{Kazi~Injamamul Haque}, {and} \bibinfo{person}{Zerrin Yumak}.} \bibinfo{year}{2023}\natexlab{}.
\newblock \showarticletitle{Facediffuser: Speech-driven 3d facial animation synthesis using diffusion}. In \bibinfo{booktitle}{\emph{Proceedings of the 16th ACM SIGGRAPH Conference on Motion, Interaction and Games}}. \bibinfo{pages}{1--11}.
\newblock


\bibitem[Sun et~al\mbox{.}(2023)]%
        {sun2023diffposetalk}
\bibfield{author}{\bibinfo{person}{Zhiyao Sun}, \bibinfo{person}{Tian Lv}, \bibinfo{person}{Sheng Ye}, \bibinfo{person}{Matthieu~Gaetan Lin}, \bibinfo{person}{Jenny Sheng}, \bibinfo{person}{Yu-Hui Wen}, \bibinfo{person}{Minjing Yu}, {and} \bibinfo{person}{Yong-jin Liu}.} \bibinfo{year}{2023}\natexlab{}.
\newblock \showarticletitle{Diffposetalk: Speech-driven stylistic 3d facial animation and head pose generation via diffusion models}.
\newblock \bibinfo{journal}{\emph{arXiv preprint arXiv:2310.00434}} (\bibinfo{year}{2023}).
\newblock


\bibitem[Taylor et~al\mbox{.}(2012)]%
        {taylor2012dynamic}
\bibfield{author}{\bibinfo{person}{Sarah~L Taylor}, \bibinfo{person}{Moshe Mahler}, \bibinfo{person}{Barry-John Theobald}, {and} \bibinfo{person}{Iain Matthews}.} \bibinfo{year}{2012}\natexlab{}.
\newblock \showarticletitle{Dynamic units of visual speech}. In \bibinfo{booktitle}{\emph{Proceedings of the 11th ACM SIGGRAPH/Eurographics conference on Computer Animation}}. \bibinfo{pages}{275--284}.
\newblock


\bibitem[Tevet et~al\mbox{.}(2022a)]%
        {tevet2022motionclip}
\bibfield{author}{\bibinfo{person}{Guy Tevet}, \bibinfo{person}{Brian Gordon}, \bibinfo{person}{Amir Hertz}, \bibinfo{person}{Amit~H Bermano}, {and} \bibinfo{person}{Daniel Cohen-Or}.} \bibinfo{year}{2022}\natexlab{a}.
\newblock \showarticletitle{Motionclip: Exposing human motion generation to clip space}. In \bibinfo{booktitle}{\emph{European Conference on Computer Vision}}. Springer, \bibinfo{pages}{358--374}.
\newblock


\bibitem[Tevet et~al\mbox{.}(2022b)]%
        {tevet2022human}
\bibfield{author}{\bibinfo{person}{Guy Tevet}, \bibinfo{person}{Sigal Raab}, \bibinfo{person}{Brian Gordon}, \bibinfo{person}{Yonatan Shafir}, \bibinfo{person}{Daniel Cohen-Or}, {and} \bibinfo{person}{Amit~H Bermano}.} \bibinfo{year}{2022}\natexlab{b}.
\newblock \showarticletitle{Human motion diffusion model}.
\newblock \bibinfo{journal}{\emph{arXiv preprint arXiv:2209.14916}} (\bibinfo{year}{2022}).
\newblock


\bibitem[Thambiraja et~al\mbox{.}(2023a)]%
        {thambiraja20233diface}
\bibfield{author}{\bibinfo{person}{Balamurugan Thambiraja}, \bibinfo{person}{Sadegh Aliakbarian}, \bibinfo{person}{Darren Cosker}, {and} \bibinfo{person}{Justus Thies}.} \bibinfo{year}{2023}\natexlab{a}.
\newblock \showarticletitle{3diface: Diffusion-based speech-driven 3d facial animation and editing}.
\newblock \bibinfo{journal}{\emph{arXiv preprint arXiv:2312.00870}} (\bibinfo{year}{2023}).
\newblock


\bibitem[Thambiraja et~al\mbox{.}(2023b)]%
        {thambiraja2023imitator}
\bibfield{author}{\bibinfo{person}{Balamurugan Thambiraja}, \bibinfo{person}{Ikhsanul Habibie}, \bibinfo{person}{Sadegh Aliakbarian}, \bibinfo{person}{Darren Cosker}, \bibinfo{person}{Christian Theobalt}, {and} \bibinfo{person}{Justus Thies}.} \bibinfo{year}{2023}\natexlab{b}.
\newblock \showarticletitle{Imitator: Personalized speech-driven 3d facial animation}. In \bibinfo{booktitle}{\emph{Proceedings of the IEEE/CVF International Conference on Computer Vision}}. \bibinfo{pages}{20621--20631}.
\newblock


\bibitem[Van Den~Oord et~al\mbox{.}(2017)]%
        {van2017neural}
\bibfield{author}{\bibinfo{person}{Aaron Van Den~Oord}, \bibinfo{person}{Oriol Vinyals}, {et~al\mbox{.}}} \bibinfo{year}{2017}\natexlab{}.
\newblock \showarticletitle{Neural discrete representation learning}.
\newblock \bibinfo{journal}{\emph{Advances in neural information processing systems}}  \bibinfo{volume}{30} (\bibinfo{year}{2017}).
\newblock


\bibitem[Vaswani et~al\mbox{.}(2017)]%
        {vaswani2017attention}
\bibfield{author}{\bibinfo{person}{Ashish Vaswani}, \bibinfo{person}{Noam Shazeer}, \bibinfo{person}{Niki Parmar}, \bibinfo{person}{Jakob Uszkoreit}, \bibinfo{person}{Llion Jones}, \bibinfo{person}{Aidan~N Gomez}, \bibinfo{person}{{\L}ukasz Kaiser}, {and} \bibinfo{person}{Illia Polosukhin}.} \bibinfo{year}{2017}\natexlab{}.
\newblock \showarticletitle{Attention is all you need}.
\newblock \bibinfo{journal}{\emph{Advances in neural information processing systems}}  \bibinfo{volume}{30} (\bibinfo{year}{2017}).
\newblock


\bibitem[Wang et~al\mbox{.}(2023a)]%
        {wang2023agentavatar}
\bibfield{author}{\bibinfo{person}{Duomin Wang}, \bibinfo{person}{Bin Dai}, \bibinfo{person}{Yu Deng}, {and} \bibinfo{person}{Baoyuan Wang}.} \bibinfo{year}{2023}\natexlab{a}.
\newblock \showarticletitle{Agentavatar: Disentangling planning, driving and rendering for photorealistic avatar agents}.
\newblock \bibinfo{journal}{\emph{arXiv preprint arXiv:2311.17465}} (\bibinfo{year}{2023}).
\newblock


\bibitem[Wang et~al\mbox{.}(2023b)]%
        {wang2023versatile}
\bibfield{author}{\bibinfo{person}{Haoyu Wang}, \bibinfo{person}{Haozhe Wu}, \bibinfo{person}{Junliang Xing}, {and} \bibinfo{person}{Jia Jia}.} \bibinfo{year}{2023}\natexlab{b}.
\newblock \showarticletitle{Versatile Face Animator: Driving Arbitrary 3D Facial Avatar in RGBD Space}. In \bibinfo{booktitle}{\emph{Proceedings of the 31st ACM International Conference on Multimedia}}. \bibinfo{pages}{7776--7784}.
\newblock


\bibitem[Wang et~al\mbox{.}(2023c)]%
        {wang2023modelscope}
\bibfield{author}{\bibinfo{person}{Jiuniu Wang}, \bibinfo{person}{Hangjie Yuan}, \bibinfo{person}{Dayou Chen}, \bibinfo{person}{Yingya Zhang}, \bibinfo{person}{Xiang Wang}, {and} \bibinfo{person}{Shiwei Zhang}.} \bibinfo{year}{2023}\natexlab{c}.
\newblock \showarticletitle{Modelscope text-to-video technical report}.
\newblock \bibinfo{journal}{\emph{arXiv preprint arXiv:2308.06571}} (\bibinfo{year}{2023}).
\newblock


\bibitem[Wang et~al\mbox{.}(2020)]%
        {wang2020mead}
\bibfield{author}{\bibinfo{person}{Kaisiyuan Wang}, \bibinfo{person}{Qianyi Wu}, \bibinfo{person}{Linsen Song}, \bibinfo{person}{Zhuoqian Yang}, \bibinfo{person}{Wayne Wu}, \bibinfo{person}{Chen Qian}, \bibinfo{person}{Ran He}, \bibinfo{person}{Yu Qiao}, {and} \bibinfo{person}{Chen~Change Loy}.} \bibinfo{year}{2020}\natexlab{}.
\newblock \showarticletitle{Mead: A large-scale audio-visual dataset for emotional talking-face generation}. In \bibinfo{booktitle}{\emph{European Conference on Computer Vision}}. Springer, \bibinfo{pages}{700--717}.
\newblock


\bibitem[Wang et~al\mbox{.}(2021)]%
        {wang20213d}
\bibfield{author}{\bibinfo{person}{Qianyun Wang}, \bibinfo{person}{Zhenfeng Fan}, {and} \bibinfo{person}{Shihong Xia}.} \bibinfo{year}{2021}\natexlab{}.
\newblock \showarticletitle{3d-talkemo: Learning to synthesize 3d emotional talking head}.
\newblock \bibinfo{journal}{\emph{arXiv preprint arXiv:2104.12051}} (\bibinfo{year}{2021}).
\newblock


\bibitem[Wu et~al\mbox{.}(2023a)]%
        {wu2023mmface4d}
\bibfield{author}{\bibinfo{person}{Haozhe Wu}, \bibinfo{person}{Jia Jia}, \bibinfo{person}{Junliang Xing}, \bibinfo{person}{Hongwei Xu}, \bibinfo{person}{Xiangyuan Wang}, {and} \bibinfo{person}{Jelo Wang}.} \bibinfo{year}{2023}\natexlab{a}.
\newblock \showarticletitle{MMFace4D: A Large-Scale Multi-Modal 4D Face Dataset for Audio-Driven 3D Face Animation}.
\newblock \bibinfo{journal}{\emph{arXiv preprint arXiv:2303.09797}} (\bibinfo{year}{2023}).
\newblock


\bibitem[Wu et~al\mbox{.}(2023d)]%
        {wu2023speech}
\bibfield{author}{\bibinfo{person}{Haozhe Wu}, \bibinfo{person}{Songtao Zhou}, \bibinfo{person}{Jia Jia}, \bibinfo{person}{Junliang Xing}, \bibinfo{person}{Qi Wen}, {and} \bibinfo{person}{Xiang Wen}.} \bibinfo{year}{2023}\natexlab{d}.
\newblock \showarticletitle{Speech-Driven 3D Face Animation with Composite and Regional Facial Movements}. In \bibinfo{booktitle}{\emph{Proceedings of the 31st ACM International Conference on Multimedia}}. \bibinfo{pages}{6822--6830}.
\newblock


\bibitem[Wu et~al\mbox{.}(2023b)]%
        {wu2023singinghead}
\bibfield{author}{\bibinfo{person}{Sijing Wu}, \bibinfo{person}{Yunhao Li}, \bibinfo{person}{Weitian Zhang}, \bibinfo{person}{Jun Jia}, \bibinfo{person}{Yucheng Zhu}, \bibinfo{person}{Yichao Yan}, {and} \bibinfo{person}{Guangtao Zhai}.} \bibinfo{year}{2023}\natexlab{b}.
\newblock \showarticletitle{SingingHead: A Large-scale 4D Dataset for Singing Head Animation}.
\newblock \bibinfo{journal}{\emph{arXiv preprint arXiv:2312.04369}} (\bibinfo{year}{2023}).
\newblock


\bibitem[Wu et~al\mbox{.}(2023c)]%
        {wu2023ganhead}
\bibfield{author}{\bibinfo{person}{Sijing Wu}, \bibinfo{person}{Yichao Yan}, \bibinfo{person}{Yunhao Li}, \bibinfo{person}{Yuhao Cheng}, \bibinfo{person}{Wenhan Zhu}, \bibinfo{person}{Ke Gao}, \bibinfo{person}{Xiaobo Li}, {and} \bibinfo{person}{Guangtao Zhai}.} \bibinfo{year}{2023}\natexlab{c}.
\newblock \showarticletitle{GANHead: Towards Generative Animatable Neural Head Avatars}. In \bibinfo{booktitle}{\emph{Proceedings of the IEEE/CVF Conference on Computer Vision and Pattern Recognition}}. \bibinfo{pages}{437--447}.
\newblock


\bibitem[Wuu et~al\mbox{.}(2022)]%
        {wuu2022multiface}
\bibfield{author}{\bibinfo{person}{Cheng-hsin Wuu}, \bibinfo{person}{Ningyuan Zheng}, \bibinfo{person}{Scott Ardisson}, \bibinfo{person}{Rohan Bali}, \bibinfo{person}{Danielle Belko}, \bibinfo{person}{Eric Brockmeyer}, \bibinfo{person}{Lucas Evans}, \bibinfo{person}{Timothy Godisart}, \bibinfo{person}{Hyowon Ha}, \bibinfo{person}{Xuhua Huang}, \bibinfo{person}{Alexander Hypes}, \bibinfo{person}{Taylor Koska}, \bibinfo{person}{Steven Krenn}, \bibinfo{person}{Stephen Lombardi}, \bibinfo{person}{Xiaomin Luo}, \bibinfo{person}{Kevyn McPhail}, \bibinfo{person}{Laura Millerschoen}, \bibinfo{person}{Michal Perdoch}, \bibinfo{person}{Mark Pitts}, \bibinfo{person}{Alexander Richard}, \bibinfo{person}{Jason Saragih}, \bibinfo{person}{Junko Saragih}, \bibinfo{person}{Takaaki Shiratori}, \bibinfo{person}{Tomas Simon}, \bibinfo{person}{Matt Stewart}, \bibinfo{person}{Autumn Trimble}, \bibinfo{person}{Xinshuo Weng}, \bibinfo{person}{David Whitewolf}, \bibinfo{person}{Chenglei Wu}, \bibinfo{person}{Shoou-I Yu}, {and}
  \bibinfo{person}{Yaser Sheikh}.} \bibinfo{year}{2022}\natexlab{}.
\newblock \showarticletitle{Multiface: A Dataset for Neural Face Rendering}. In \bibinfo{booktitle}{\emph{arXiv}}.
\newblock
\urldef\tempurl%
\url{https://doi.org/10.48550/ARXIV.2207.11243}
\showDOI{\tempurl}


\bibitem[Xia et~al\mbox{.}(2021)]%
        {xia2021tedigan}
\bibfield{author}{\bibinfo{person}{Weihao Xia}, \bibinfo{person}{Yujiu Yang}, \bibinfo{person}{Jing-Hao Xue}, {and} \bibinfo{person}{Baoyuan Wu}.} \bibinfo{year}{2021}\natexlab{}.
\newblock \showarticletitle{Tedigan: Text-guided diverse face image generation and manipulation}. In \bibinfo{booktitle}{\emph{Proceedings of the IEEE/CVF conference on computer vision and pattern recognition}}. \bibinfo{pages}{2256--2265}.
\newblock


\bibitem[Xing et~al\mbox{.}(2023)]%
        {xing2023codetalker}
\bibfield{author}{\bibinfo{person}{Jinbo Xing}, \bibinfo{person}{Menghan Xia}, \bibinfo{person}{Yuechen Zhang}, \bibinfo{person}{Xiaodong Cun}, \bibinfo{person}{Jue Wang}, {and} \bibinfo{person}{Tien-Tsin Wong}.} \bibinfo{year}{2023}\natexlab{}.
\newblock \showarticletitle{Codetalker: Speech-driven 3d facial animation with discrete motion prior}. In \bibinfo{booktitle}{\emph{Proceedings of the IEEE/CVF Conference on Computer Vision and Pattern Recognition}}. \bibinfo{pages}{12780--12790}.
\newblock


\bibitem[Xu et~al\mbox{.}(2023)]%
        {xu2023inter}
\bibfield{author}{\bibinfo{person}{Liang Xu}, \bibinfo{person}{Xintao Lv}, \bibinfo{person}{Yichao Yan}, \bibinfo{person}{Xin Jin}, \bibinfo{person}{Shuwen Wu}, \bibinfo{person}{Congsheng Xu}, \bibinfo{person}{Yifan Liu}, \bibinfo{person}{Yizhou Zhou}, \bibinfo{person}{Fengyun Rao}, \bibinfo{person}{Xingdong Sheng}, {et~al\mbox{.}}} \bibinfo{year}{2023}\natexlab{}.
\newblock \showarticletitle{Inter-X: Towards Versatile Human-Human Interaction Analysis}.
\newblock \bibinfo{journal}{\emph{arXiv preprint arXiv:2312.16051}} (\bibinfo{year}{2023}).
\newblock


\bibitem[Xu et~al\mbox{.}(2013)]%
        {xu2013practical}
\bibfield{author}{\bibinfo{person}{Yuyu Xu}, \bibinfo{person}{Andrew~W Feng}, \bibinfo{person}{Stacy Marsella}, {and} \bibinfo{person}{Ari Shapiro}.} \bibinfo{year}{2013}\natexlab{}.
\newblock \showarticletitle{A practical and configurable lip sync method for games}.
\newblock In \bibinfo{booktitle}{\emph{Proceedings of Motion on Games}}. \bibinfo{pages}{131--140}.
\newblock


\bibitem[Yang et~al\mbox{.}(2023a)]%
        {yang2023probabilistic}
\bibfield{author}{\bibinfo{person}{Karren~D Yang}, \bibinfo{person}{Anurag Ranjan}, \bibinfo{person}{Jen-Hao~Rick Chang}, \bibinfo{person}{Raviteja Vemulapalli}, {and} \bibinfo{person}{Oncel Tuzel}.} \bibinfo{year}{2023}\natexlab{a}.
\newblock \showarticletitle{Probabilistic Speech-Driven 3D Facial Motion Synthesis: New Benchmarks, Methods, and Applications}.
\newblock \bibinfo{journal}{\emph{arXiv preprint arXiv:2311.18168}} (\bibinfo{year}{2023}).
\newblock


\bibitem[Yang et~al\mbox{.}(2023b)]%
        {yang2023synthesizing}
\bibfield{author}{\bibinfo{person}{Zhao Yang}, \bibinfo{person}{Bing Su}, {and} \bibinfo{person}{Ji-Rong Wen}.} \bibinfo{year}{2023}\natexlab{b}.
\newblock \showarticletitle{Synthesizing Long-Term Human Motions with Diffusion Models via Coherent Sampling}. In \bibinfo{booktitle}{\emph{Proceedings of the 31st ACM International Conference on Multimedia}}. \bibinfo{pages}{3954--3964}.
\newblock


\bibitem[Yu et~al\mbox{.}(2022)]%
        {yu2022scaling}
\bibfield{author}{\bibinfo{person}{Jiahui Yu}, \bibinfo{person}{Yuanzhong Xu}, \bibinfo{person}{Jing~Yu Koh}, \bibinfo{person}{Thang Luong}, \bibinfo{person}{Gunjan Baid}, \bibinfo{person}{Zirui Wang}, \bibinfo{person}{Vijay Vasudevan}, \bibinfo{person}{Alexander Ku}, \bibinfo{person}{Yinfei Yang}, \bibinfo{person}{Burcu~Karagol Ayan}, {et~al\mbox{.}}} \bibinfo{year}{2022}\natexlab{}.
\newblock \showarticletitle{Scaling autoregressive models for content-rich text-to-image generation}.
\newblock \bibinfo{journal}{\emph{arXiv preprint arXiv:2206.10789}} (\bibinfo{year}{2022}).
\newblock


\bibitem[Yu et~al\mbox{.}(2023)]%
        {yu2023celebv}
\bibfield{author}{\bibinfo{person}{Jianhui Yu}, \bibinfo{person}{Hao Zhu}, \bibinfo{person}{Liming Jiang}, \bibinfo{person}{Chen~Change Loy}, \bibinfo{person}{Weidong Cai}, {and} \bibinfo{person}{Wayne Wu}.} \bibinfo{year}{2023}\natexlab{}.
\newblock \showarticletitle{CelebV-Text: A Large-Scale Facial Text-Video Dataset}. In \bibinfo{booktitle}{\emph{Proceedings of the IEEE/CVF Conference on Computer Vision and Pattern Recognition}}. \bibinfo{pages}{14805--14814}.
\newblock


\bibitem[Zhai et~al\mbox{.}(2023)]%
        {zhai2023language}
\bibfield{author}{\bibinfo{person}{Yuanhao Zhai}, \bibinfo{person}{Mingzhen Huang}, \bibinfo{person}{Tianyu Luan}, \bibinfo{person}{Lu Dong}, \bibinfo{person}{Ifeoma Nwogu}, \bibinfo{person}{Siwei Lyu}, \bibinfo{person}{David Doermann}, {and} \bibinfo{person}{Junsong Yuan}.} \bibinfo{year}{2023}\natexlab{}.
\newblock \showarticletitle{Language-guided Human Motion Synthesis with Atomic Actions}. In \bibinfo{booktitle}{\emph{Proceedings of the 31st ACM International Conference on Multimedia}}. \bibinfo{pages}{5262--5271}.
\newblock


\bibitem[Zhang et~al\mbox{.}(2023a)]%
        {zhang2023show}
\bibfield{author}{\bibinfo{person}{David~Junhao Zhang}, \bibinfo{person}{Jay~Zhangjie Wu}, \bibinfo{person}{Jia-Wei Liu}, \bibinfo{person}{Rui Zhao}, \bibinfo{person}{Lingmin Ran}, \bibinfo{person}{Yuchao Gu}, \bibinfo{person}{Difei Gao}, {and} \bibinfo{person}{Mike~Zheng Shou}.} \bibinfo{year}{2023}\natexlab{a}.
\newblock \showarticletitle{Show-1: Marrying pixel and latent diffusion models for text-to-video generation}.
\newblock \bibinfo{journal}{\emph{arXiv preprint arXiv:2309.15818}} (\bibinfo{year}{2023}).
\newblock


\bibitem[Zhang et~al\mbox{.}(2023b)]%
        {zhang2023generating}
\bibfield{author}{\bibinfo{person}{Jianrong Zhang}, \bibinfo{person}{Yangsong Zhang}, \bibinfo{person}{Xiaodong Cun}, \bibinfo{person}{Shaoli Huang}, \bibinfo{person}{Yong Zhang}, \bibinfo{person}{Hongwei Zhao}, \bibinfo{person}{Hongtao Lu}, {and} \bibinfo{person}{Xi Shen}.} \bibinfo{year}{2023}\natexlab{b}.
\newblock \showarticletitle{T2M-GPT: Generating Human Motion from Textual Descriptions with Discrete Representations}. In \bibinfo{booktitle}{\emph{Proceedings of the IEEE/CVF Conference on Computer Vision and Pattern Recognition (CVPR)}}.
\newblock


\bibitem[Zhang et~al\mbox{.}(2023c)]%
        {zhang2023t2m}
\bibfield{author}{\bibinfo{person}{Jianrong Zhang}, \bibinfo{person}{Yangsong Zhang}, \bibinfo{person}{Xiaodong Cun}, \bibinfo{person}{Shaoli Huang}, \bibinfo{person}{Yong Zhang}, \bibinfo{person}{Hongwei Zhao}, \bibinfo{person}{Hongtao Lu}, {and} \bibinfo{person}{Xi Shen}.} \bibinfo{year}{2023}\natexlab{c}.
\newblock \showarticletitle{T2m-gpt: Generating human motion from textual descriptions with discrete representations}.
\newblock \bibinfo{journal}{\emph{arXiv preprint arXiv:2301.06052}} (\bibinfo{year}{2023}).
\newblock


\bibitem[Zhang et~al\mbox{.}(2024)]%
        {zhang2024motiondiffuse}
\bibfield{author}{\bibinfo{person}{Mingyuan Zhang}, \bibinfo{person}{Zhongang Cai}, \bibinfo{person}{Liang Pan}, \bibinfo{person}{Fangzhou Hong}, \bibinfo{person}{Xinying Guo}, \bibinfo{person}{Lei Yang}, {and} \bibinfo{person}{Ziwei Liu}.} \bibinfo{year}{2024}\natexlab{}.
\newblock \showarticletitle{Motiondiffuse: Text-driven human motion generation with diffusion model}.
\newblock \bibinfo{journal}{\emph{IEEE Transactions on Pattern Analysis and Machine Intelligence}} (\bibinfo{year}{2024}).
\newblock


\bibitem[Zhang et~al\mbox{.}(2014)]%
        {zhang2014bp4d}
\bibfield{author}{\bibinfo{person}{Xing Zhang}, \bibinfo{person}{Lijun Yin}, \bibinfo{person}{Jeffrey~F Cohn}, \bibinfo{person}{Shaun Canavan}, \bibinfo{person}{Michael Reale}, \bibinfo{person}{Andy Horowitz}, \bibinfo{person}{Peng Liu}, {and} \bibinfo{person}{Jeffrey~M Girard}.} \bibinfo{year}{2014}\natexlab{}.
\newblock \showarticletitle{Bp4d-spontaneous: a high-resolution spontaneous 3d dynamic facial expression database}.
\newblock \bibinfo{journal}{\emph{Image and Vision Computing}} \bibinfo{volume}{32}, \bibinfo{number}{10} (\bibinfo{year}{2014}), \bibinfo{pages}{692--706}.
\newblock


\bibitem[Zhang et~al\mbox{.}(2016)]%
        {zhang2016multimodal}
\bibfield{author}{\bibinfo{person}{Zheng Zhang}, \bibinfo{person}{Jeff~M Girard}, \bibinfo{person}{Yue Wu}, \bibinfo{person}{Xing Zhang}, \bibinfo{person}{Peng Liu}, \bibinfo{person}{Umur Ciftci}, \bibinfo{person}{Shaun Canavan}, \bibinfo{person}{Michael Reale}, \bibinfo{person}{Andy Horowitz}, \bibinfo{person}{Huiyuan Yang}, {et~al\mbox{.}}} \bibinfo{year}{2016}\natexlab{}.
\newblock \showarticletitle{Multimodal spontaneous emotion corpus for human behavior analysis}. In \bibinfo{booktitle}{\emph{Proceedings of the IEEE conference on computer vision and pattern recognition}}. \bibinfo{pages}{3438--3446}.
\newblock


\bibitem[Zhao et~al\mbox{.}(2024)]%
        {zhao2024media2face}
\bibfield{author}{\bibinfo{person}{Qingcheng Zhao}, \bibinfo{person}{Pengyu Long}, \bibinfo{person}{Qixuan Zhang}, \bibinfo{person}{Dafei Qin}, \bibinfo{person}{Han Liang}, \bibinfo{person}{Longwen Zhang}, \bibinfo{person}{Yingliang Zhang}, \bibinfo{person}{Jingyi Yu}, {and} \bibinfo{person}{Lan Xu}.} \bibinfo{year}{2024}\natexlab{}.
\newblock \showarticletitle{Media2face: Co-speech facial animation generation with multi-modality guidance}.
\newblock \bibinfo{journal}{\emph{arXiv preprint arXiv:2401.15687}} (\bibinfo{year}{2024}).
\newblock


\bibitem[Zheng et~al\mbox{.}(2022)]%
        {zheng2022avatar}
\bibfield{author}{\bibinfo{person}{Yufeng Zheng}, \bibinfo{person}{Victoria~Fern{\'a}ndez Abrevaya}, \bibinfo{person}{Marcel~C B{\"u}hler}, \bibinfo{person}{Xu Chen}, \bibinfo{person}{Michael~J Black}, {and} \bibinfo{person}{Otmar Hilliges}.} \bibinfo{year}{2022}\natexlab{}.
\newblock \showarticletitle{Im avatar: Implicit morphable head avatars from videos}. In \bibinfo{booktitle}{\emph{Proceedings of the IEEE/CVF Conference on Computer Vision and Pattern Recognition}}. \bibinfo{pages}{13545--13555}.
\newblock


\bibitem[Zhong et~al\mbox{.}(2024)]%
        {zhong2024expclip}
\bibfield{author}{\bibinfo{person}{Yicheng Zhong}, \bibinfo{person}{Huawei Wei}, \bibinfo{person}{Peiji Yang}, {and} \bibinfo{person}{Zhisheng Wang}.} \bibinfo{year}{2024}\natexlab{}.
\newblock \showarticletitle{ExpCLIP: Bridging Text and Facial Expressions via Semantic Alignment}. In \bibinfo{booktitle}{\emph{Proceedings of the AAAI Conference on Artificial Intelligence}}, Vol.~\bibinfo{volume}{38}. \bibinfo{pages}{7614--7622}.
\newblock


\bibitem[Zhu et~al\mbox{.}(2022)]%
        {zhu2022celebv}
\bibfield{author}{\bibinfo{person}{Hao Zhu}, \bibinfo{person}{Wayne Wu}, \bibinfo{person}{Wentao Zhu}, \bibinfo{person}{Liming Jiang}, \bibinfo{person}{Siwei Tang}, \bibinfo{person}{Li Zhang}, \bibinfo{person}{Ziwei Liu}, {and} \bibinfo{person}{Chen~Change Loy}.} \bibinfo{year}{2022}\natexlab{}.
\newblock \showarticletitle{CelebV-HQ: A large-scale video facial attributes dataset}. In \bibinfo{booktitle}{\emph{European conference on computer vision}}. Springer, \bibinfo{pages}{650--667}.
\newblock


\bibitem[Zou et~al\mbox{.}(2023)]%
        {zou20234d}
\bibfield{author}{\bibinfo{person}{Kaifeng Zou}, \bibinfo{person}{Sylvain Faisan}, \bibinfo{person}{Boyang Yu}, \bibinfo{person}{S{\'e}bastien Valette}, {and} \bibinfo{person}{Hyewon Seo}.} \bibinfo{year}{2023}\natexlab{}.
\newblock \showarticletitle{4D Facial Expression Diffusion Model}.
\newblock \bibinfo{journal}{\emph{arXiv preprint arXiv:2303.16611}} (\bibinfo{year}{2023}).
\newblock


\end{thebibliography}


\appendix
\clearpage
\renewcommand{\thetable}{\Alph{table}}
\renewcommand{\thefigure}{\Alph{figure}}
\setcounter{table}{0}
\setcounter{figure}{0}

\twocolumn[
\begin{center}
    \huge\bfseries Supplementary Material
    \vspace{2mm}
\end{center}
]

In the supplementary document, we first provide more details about MMHead dataset and MM2Face method in Section \ref{sec:dataset_sup} and Section \ref{sec:method}, respectively. 
Then, we show additional experimental results in Section \ref{sec:exp}. 
Finally, the limitations and broader impacts are discussed in Section \ref{sec:limit}. 
Please also refer to the accompanying video for more intuitive results.

\section{MMHead Dataset}
\label{sec:dataset_sup}

In this section, we first provide more details about how we obtain high precision 3D facial motion sequences and hierarchical text annotations from portrait videos, and then conduct comprehensive statistical analysis of the proposed MMHead dataset.

\subsection{3D Facial Motion from Portrait Videos}
We use sequences of expression and pose parameters of FLAME \cite{li2017learning} to represent 3D facial motion. To obtain FLAME parameters from portrait videos, we utilize a SOTA monocular 3D face reconstruction method, EMOCA \cite{danvevcek2022emoca,DECA:Siggraph2021,filntisis2022visual}, to estimate FLAME parameters frame by frame. Then we optimize the obtained FLAME parameters to improve the quality of 3D facial motion.
Specifically, we first calculate the mean value of shape, pose, expression, and bounding box size over time respectively, and then calculate $L_2$ distance of each value to the corresponding mean value and treat points whose distances are outside the interval $[Q_1-s\times IQR, Q_3+s\times IQR]$ as outliers and replace them, where $IQR=Q_3-Q_1$, $Q_1$ and $Q_3$ denote lower quartile and upper quartile respectively, $s$ is the scale factor.
Then, we optimize the expression and pose parameters by minimizing the following loss function using Adam optimizer \cite{kingma2014adam}:
\begin{equation}
\mathcal{L} = \left \| (\boldsymbol{\psi}_{2:T} -\boldsymbol{\psi}_{1:T-1} ) \right \|_2^2 +
\lambda \left \| (\boldsymbol{\theta}_{2:T} -\boldsymbol{\theta}_{1:T-1} ) \right \|_2^2,
\label{eq:1}
\end{equation}
where, $\boldsymbol{\psi}$ and $\boldsymbol{\theta}$ denote the FLAME expression and pose parameters respectively, $T$ is the length of the facial motion sequence, $\lambda$ is the balance weight.

\begin{figure}[h]
\centering
\includegraphics[scale=0.177]{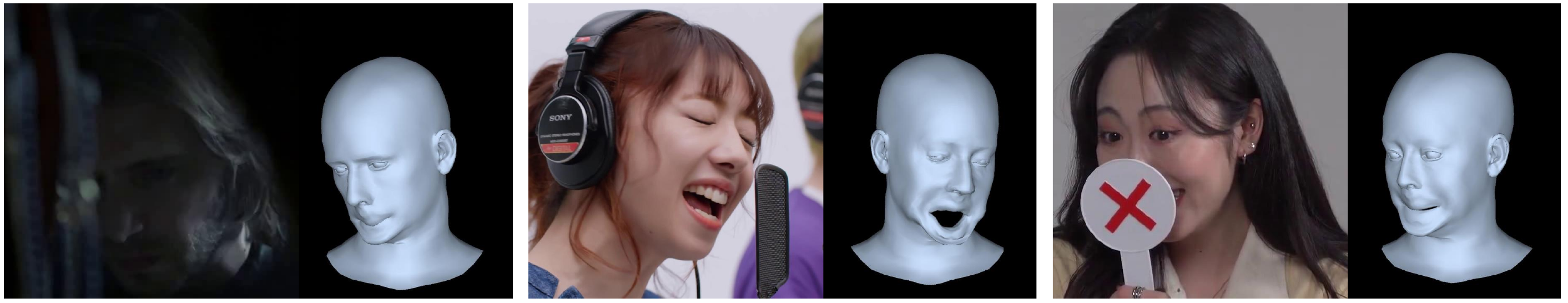}
\caption{We remove such failed cases to ensure the quality of 3D facial motion.}
\label{fig:bad_recon}
\end{figure}

\begin{figure*}[h]
\centering
\includegraphics[scale=0.4]{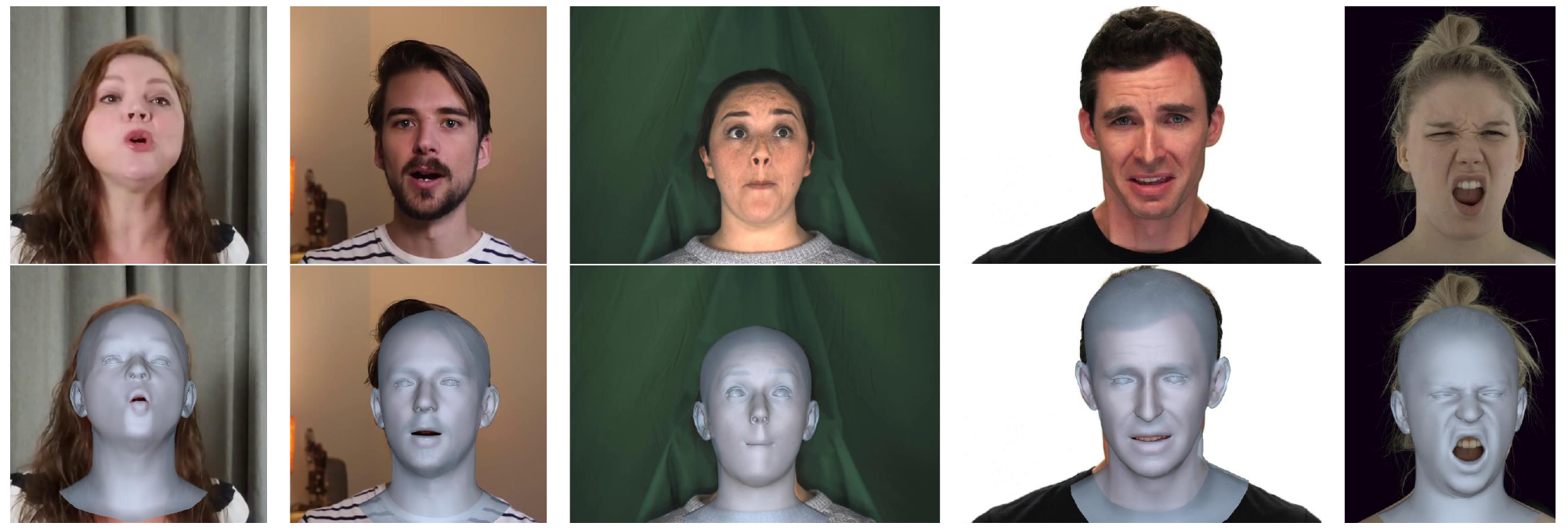}
\caption{Key frames of 3D facial motion reconstructed from portrait videos, which demonstrates the high precision of constructing 3D facial animation dataset by monocular 3D face reconstruction.}
\label{fig:good_recon}
\end{figure*}

Although the optimization step can remove the outliers and smooth the 3D facial motion sequences, motion sequences in which vast majority of frames cannot be correctly reconstructed are difficult to find and remove. To this end, we manually check the dataset and delete the bad reconstruction results, we summarized some common failure reconstruction results in Fig. \ref{fig:bad_recon}. Finally, we obtain various great reconstruction results as shown in Fig. \ref{fig:good_recon}

\subsection{Text Annotation}
\begin{figure*}
\centering
\includegraphics[scale=0.4]{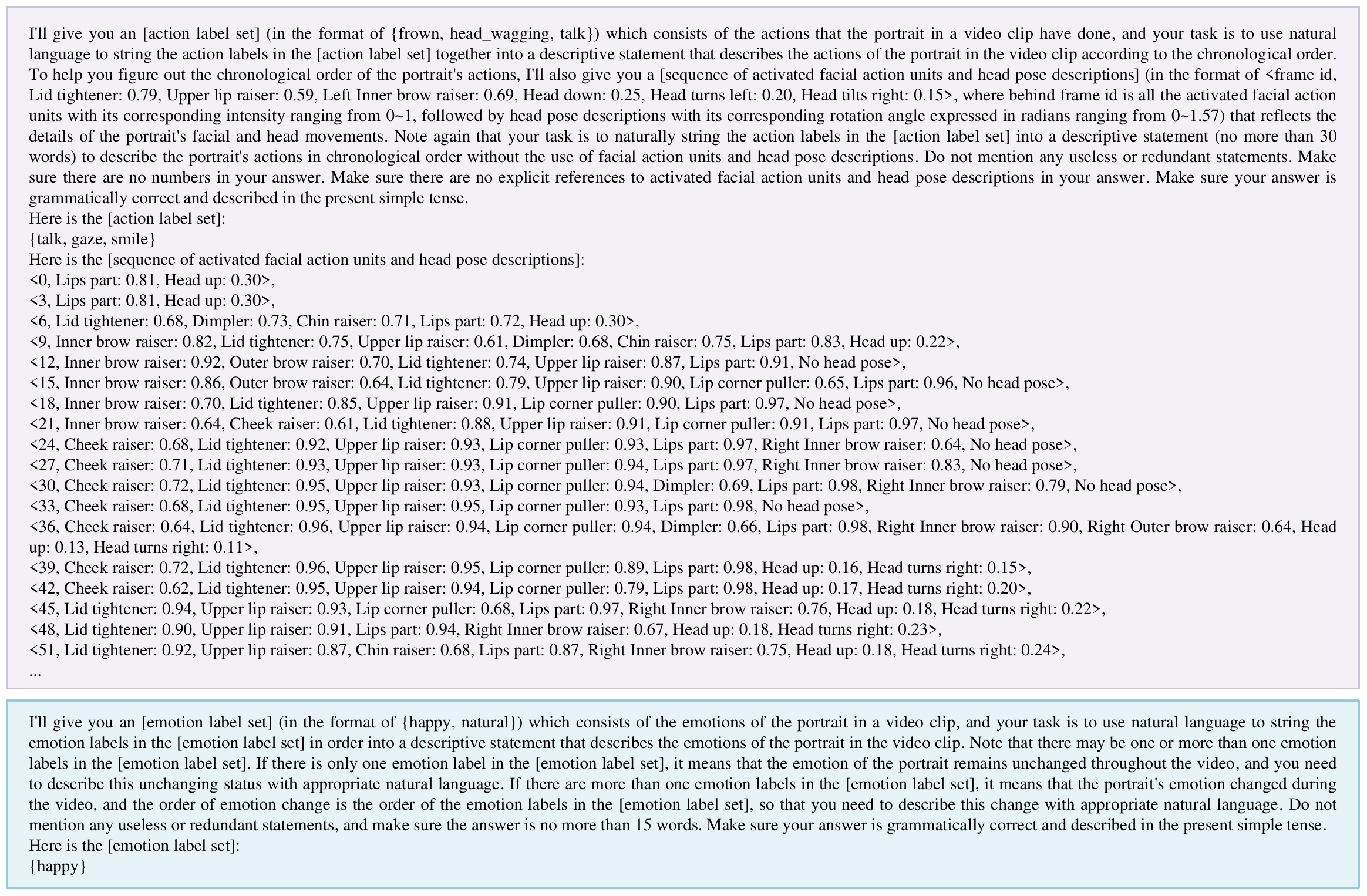}
\caption{Prompts used to annotate abstract action and emotion descriptions.}
\label{fig:prompt_abs}
\end{figure*}

\begin{figure*}
\centering
\includegraphics[scale=0.4]{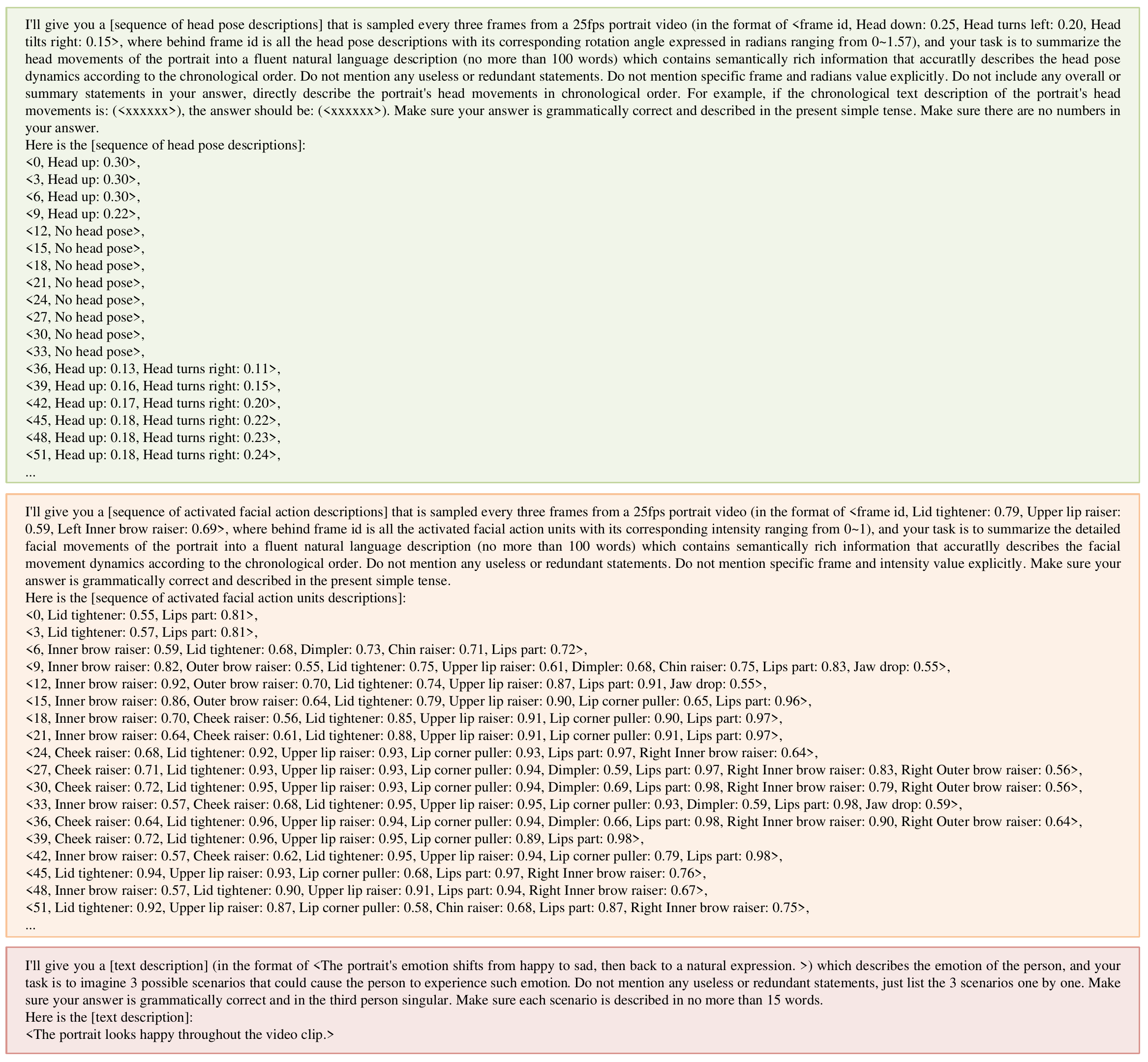}
\caption{Prompts used to annotate fine-grained head and facial movements (\textit{i.e.}, head pose and expression) descriptions, and possible emotion scenarios.}
\label{fig:prompt_detail}
\end{figure*}

As described in the main paper, we design five different prompts for annotating abstract action and emotion descriptions, fine-grained head and facial movements (\textit{i.e.}, head pose and expression) descriptions, and emotion scenarios, respectively. The prompts we used is shown in Fig. \ref{fig:prompt_abs} and Fig. \ref{fig:prompt_detail}. 
Specifically, we follow \cite{wang2023agentavatar} and use \cite{luo2022learning} to detect 41 facial action units (AU) frame by frame. As for head pose, we define 6 head pose descriptions according to the rotation vector $\boldsymbol{r}=(r_x,r_y,r_z)$ of the FLAME neck joint as illustrated in Tab. \ref{table:head_pose}.

\begin{table}[h]
\centering
\begin{tabular}{c|l}
\toprule
\textbf{Rotation Range}  & \textbf{Description} \\ 
\midrule

$r_x<0$  & Head up \\
$r_x>0$  & Head down \\
$r_y<0$  & Head turns right \\
$r_y>0$  & Head turns left \\
$r_z<0$  & Head tilts left \\
$r_z>0$  & Head tilts right \\

\bottomrule
\end{tabular}
\caption{Head pose descriptions according to the rotation vector $\boldsymbol{r}=(r_x,r_y,r_z)$ of the FLAME neck joint.}
\label{table:head_pose}
\end{table}

\subsection{Dataset Statistics}

We conduct data analysis of MMHead dataset. We gradually analyze the motion duration distribution, action distribution, emotion distribution and head pose distribution.

\begin{figure}
\centering
\includegraphics[scale=0.25]{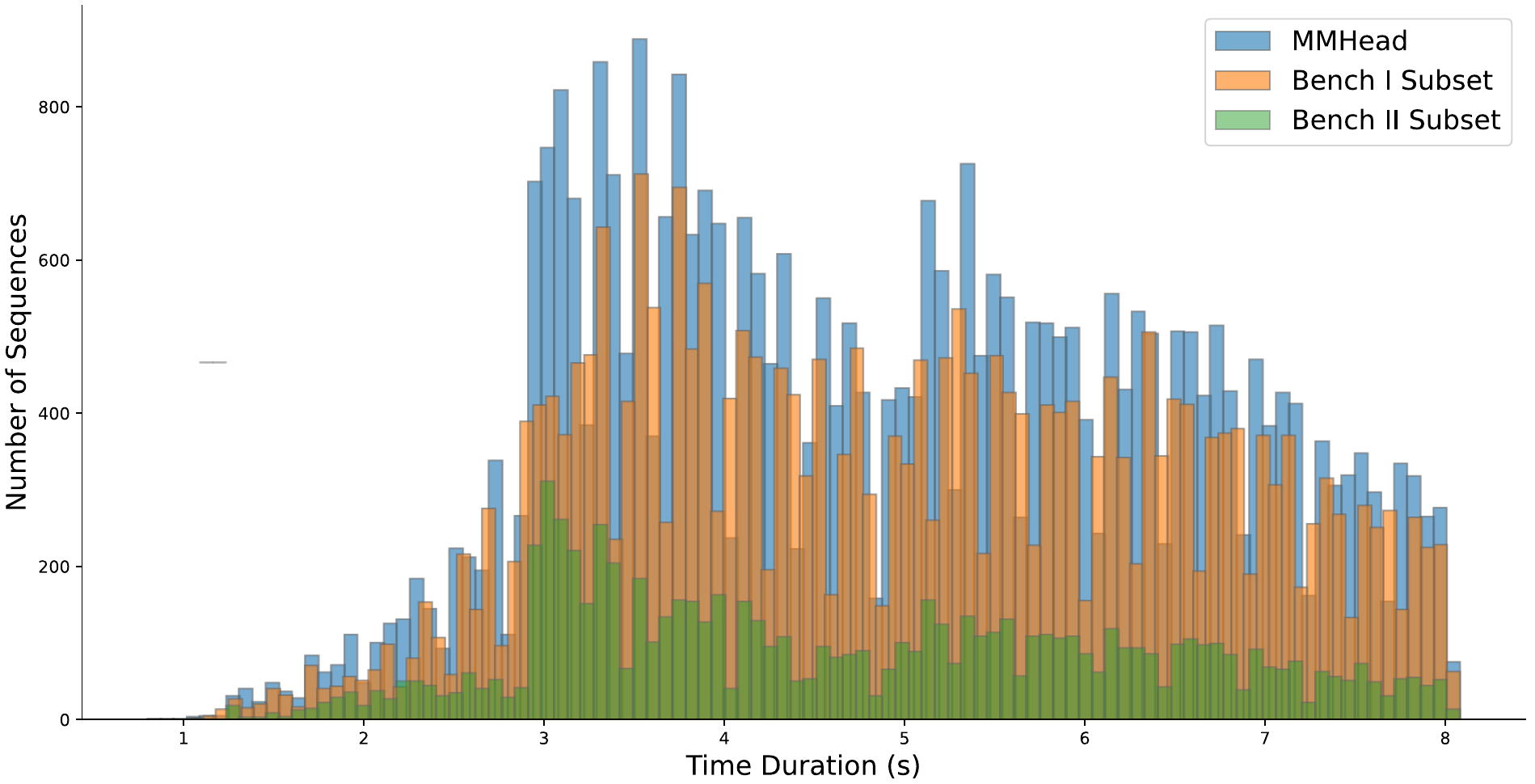}
\caption{Histograms of facial motion time duration of entire MMHead dataset, benchmark I subset, and benchmark II subset, respectively.}
\label{fig:dur_stat}
\end{figure}

\noindent\textbf{Motion Duration Distribution.} The duration distribution is summarized in Fig. \ref{fig:dur_stat}. MMHead contains diverse motions which cover a wide range of duration times (1 seconds to 8 seconds). The similar distributions of Bench I subset and Bench II subset also demonstrate that our talking animations and 3D facial motions both contains various motions.

\begin{figure}
\centering
\includegraphics[scale=0.47]{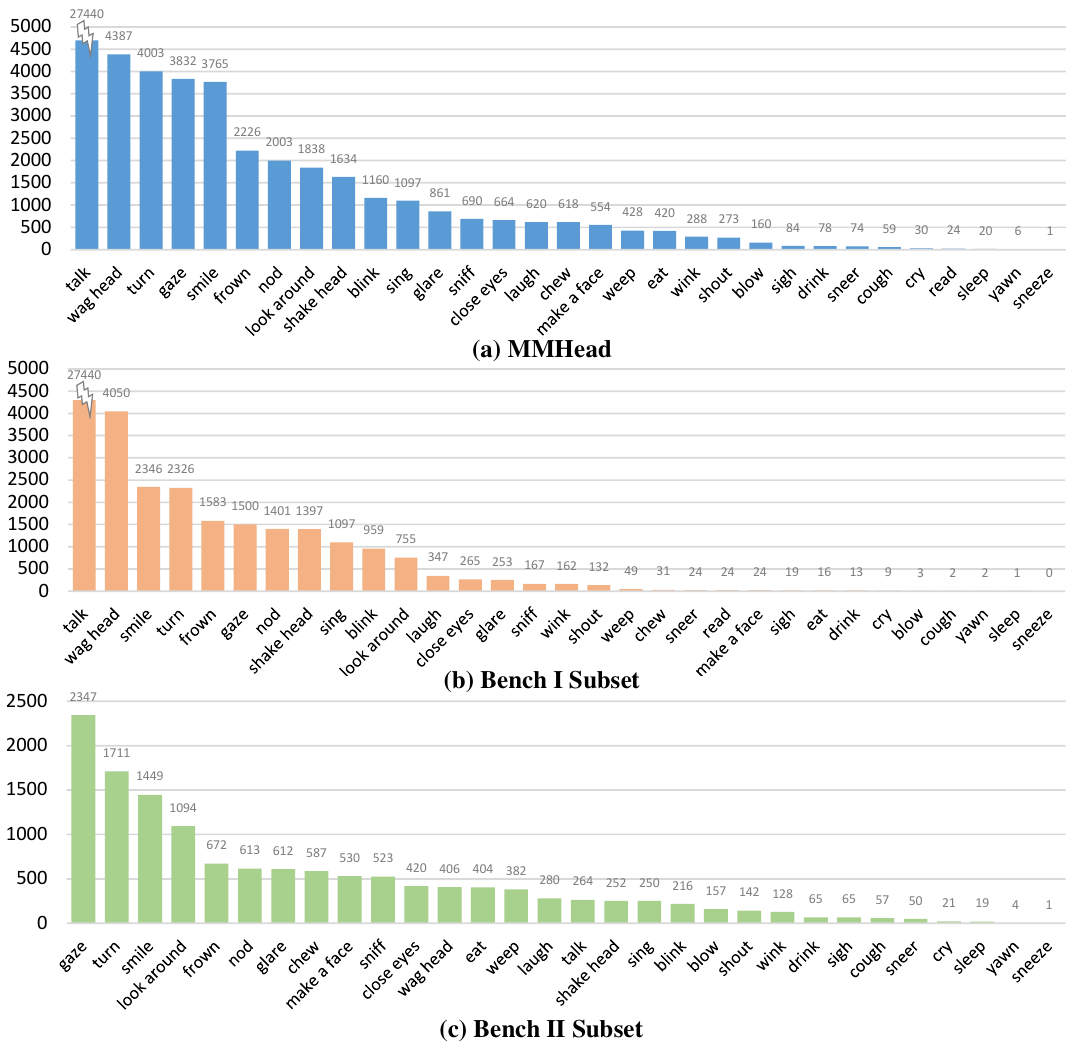}
\caption{Distributions of all actions that occur in the entire MMHead dataset, benchmark I subset, and benchmark II subset, respectively.}
\label{fig:act_stat}
\end{figure}

\noindent\textbf{Action Distribution.} The action distribution indicates the distribution of action words extracted from abstract action descriptions in our dataset, reflecting the diversity of our abstract action text descriptions. The results are summarized in Fig. \ref{fig:act_stat}. We can observe that (1) "talk" is the word with the highest frequency in MMHead dataset because most of our data are collected from talking head datasets. (2) Compared to the action distributions in MMHead and Bench I datasets, the distribution in Bench II subset is quite different, which contains many high frequency novel actions such as "make a face", "eat", "chew", "weep" etc. This demonstrate that 3D facial motion is quite different from 3D talking head because benchmark II are mainly focusing on facial expression motion generation instead of talking motions with corresponding audios.

\begin{figure}
\centering
\includegraphics[scale=0.25]{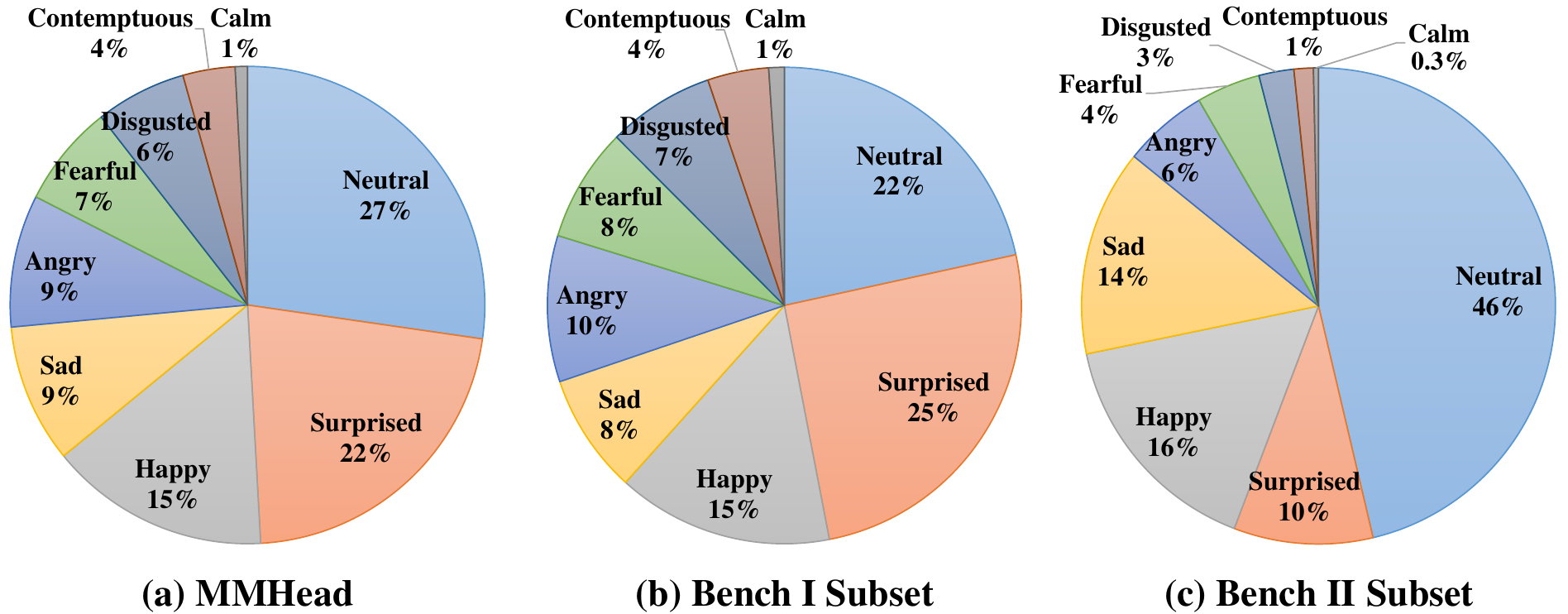}
\caption{Pie charts of emotion categories of entire MMHead dataset, benchmark I subset, and benchmark II subset, respectively.}
\label{fig:emo_stat}
\end{figure}

\noindent\textbf{Emotion Distribution.} The Fig. \ref{fig:emo_stat} reflects the diverse emotion distribution in our MMHead dataset.

\begin{figure}
\centering
\includegraphics[scale=0.43]{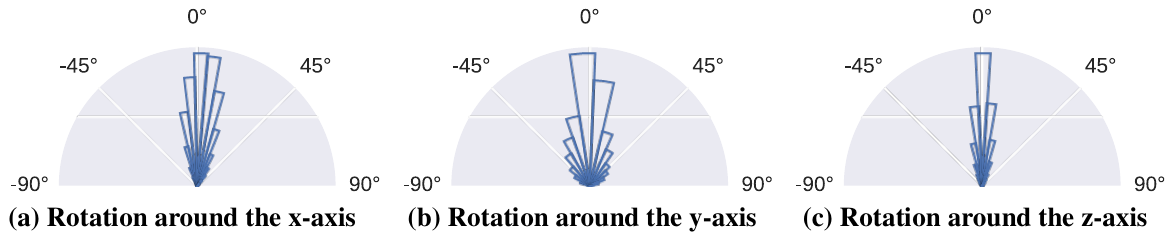}
\caption{Histograms of the head rotation around x-axis (\textit{i.e.,} up/down), y-axis (\textit{i.e.,} turn left/right), and z-axis (\textit{i.e.,} tilt left/right), respectively.}
\label{fig:pose_stat}
\end{figure}

\noindent\textbf{Head Pose Distribution.} In addition, the
distribution of head pose indicates the diversity of the dataset in head movements. As shown in Fig. \ref{fig:pose_stat}, MMHead has a uniform distribution, which indicates that the motions have diverse head movements.




\subsection{Evaluation of MMHead Dataset Quality}
\label{pipeline_eval}

\noindent\textbf{Text-to-3D Facial Motion Alignment Evaluation.} We perform a text-to-motion retrieval task to evaluate the text-to-3D facial motion alignment degree in MMHead dataset. Concretely, we select TMR \cite{petrovich2023tmr} which is a state-of-the-art retrieval method in 3D human motion area as the baseline method, and calculate the performance on both text-induced 3D talking head animation (benchmark I) and text-to-3d facial motion generation (benchmark II) subsets. We follow the common protocol \cite{guo2022generating, xu2023inter, lin2024motion, liang2024intergen} by using the R-Precision which calculates the average top-1 to top-3 text-to-motion retrieval accuracy in each data batch as evaluation metrics.

\begin{figure}
\centering
\includegraphics[scale=0.19]{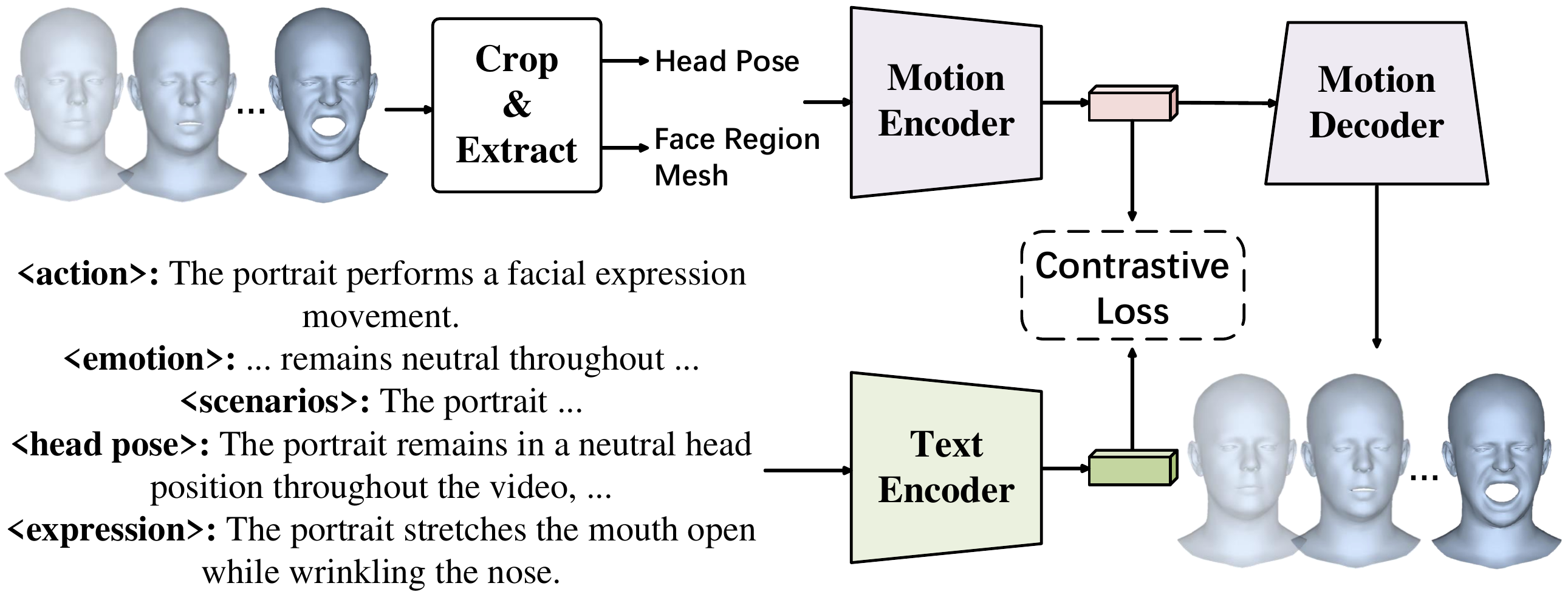}
\caption{Overview of text-to-facial motion retrieval method.}
\label{fig:tmr_text}
\end{figure}

Our detailed text-to-3D facial motion retrieval method is shown in Fig. \ref{fig:tmr_text}. The fine-grained texts from different perspectives are concatenated together and fed into a transformer-based text encoder. The 3D facial motions are cropped and extracted to masked face region mesh sequences and 3D head poses, then are fed into a transformer-based motion encoder.

The results are shown in Table. \ref{table:t2m_align}, we compare our text-to-motion retrieval accuracy with other text-to-motion datasets. We can observe that our MMHead dataset subsets achieve the highest accuracy compared to other published text-to-3D human motion datasets \cite{lin2024motion, xu2023inter, guo2022generating, liang2024intergen}, demonstrating that our text descriptions are accurate enough to conduct multi-modal generation tasks. 

\noindent\textbf{Audio-to-3D Facial Motion Alignment Evaluation.} 
Similarly, we slightly modify the text-to-motion retrieval method in Fig. \ref{fig:tmr_text} by changing the text encoder to a Wav2Vec \cite{baevski2020wav2vec} audio encoder, and removing the 3D head pose as motion inputs. After conducting the experiments, the Top-1 to Top-3 R-Precision accuracy is 0.885, 0.942, and 0.962, demonstrating the great audio-3D face motion matching performance.

\begin{table} 
\centering
\begin{tabular}{lccc}
\toprule
Dataset   & R$@$1 \,$\uparrow$ & R$@$2 \,$\uparrow$ & R$@$3 \,$\uparrow$ \\ 
\midrule

HumanML3D \cite{guo2022generating} &    0.511       &  0.703    &   0.797   \\

Motion-X \cite{lin2024motion} & 0.573         & 0.765    &  0.850     \\

Inter-X \cite{xu2023inter} & 0.429       & 0.626    &  0.736     \\

Inter-human \cite{liang2024intergen} & 0.452         & 0.610    &  0.701     \\

\rowcolor[gray]{.92}
MMHead bench I subset & \textbf{0.843}  & \textbf{0.941}  &  \textbf{0.970}   \\
\rowcolor[gray]{.92}

MMHead bench II subset & \textbf{0.678}     & \textbf{0.842}   &  \textbf{0.904}     \\

\bottomrule
\end{tabular}
\caption{\textbf{Text-to-3D facial motion retrieval performance against other text-to-3D motion datasets.} }
\label{table:t2m_align}
\end{table}

\vspace{5mm}
\section{MM2Face Method}
\label{sec:method}

\subsection{Network Architecture}
Our MM2Face method consists of a motion encoder $\mathcal{E}$, a motion decoder $\mathcal{D}$, our MM2Face transformer model $\mathcal{G}$. $\mathcal{E}$ and $\mathcal{D}$ are trained in stage 1, and $\mathcal{G}$ is trained in stage 2.

Concretely, the motion encoder $\mathcal{E}$ is comprised of ResNet based 1D CNN blocks, each block contains 1D CNN layers and normalization layers, we adopt relu function as activation function. Similarly, the motion decoder $\mathcal{D}$ is comprised of ResNet \cite{he2016deep} based 1D CNN blocks with upsample layers. We upsample the network features with nearest neighbor search strategy. The transformer model $\mathcal{G}$ is a typical layers transformer decoder models. We select Wav2Vec 2.0 \cite{baevski2020wav2vec} as our audio encoder and freeze the parameters of its feature extractor. We select pre-trained distilbert \cite{sanh2019distilbert} model as our text encoder.

\subsection{Implementation Details}
We use PyTorch to implement our MM2Face model, and utilize Adam optimizer \cite{kingma2014adam} with $[\beta_1, \beta_2]=[0.5, 0.999]$ for training. All the audio signals are normalized by Wav2Vec2.0 \cite{baevski2020wav2vec} audio processor before training. Training the stage I model takes about 10 hours on 1 Nvidia 3090 GPUs, stage II takes about 26 hours on 1 Nvidia 3090 GPUs.


\vspace{5mm}
\section{Additional Experimental Results}
\label{sec:exp}

\subsection{Additional Ablation Study} 

We also provides the ablation results of all the comparison methods to clarify the performance difference between results induced by text and those driven solely by audio in Table. \ref{table:bench1}. The results show the degradation in text matching performance of all methods and therefore demonstrates the significance of text input.

\begin{table}[h]
    \centering
    \resizebox{\linewidth}{!}{ 
    \begin{tabular}{c|l|cccccccc}
    \toprule
    & \multirow{2}{*}{Method}  & \multicolumn{3}{c}{Text R-Precision $\uparrow$} & \multirow{2}{*}{FID $\downarrow$} & \multirow{2}{*}{Audio-Match $\downarrow$} & \multirow{2}{*}{Diversity $\rightarrow$} & \multirow{2}{*}{LVE$\downarrow$} & \multirow{2}{*}{FVE $\downarrow$} \\

    \cline{3-5}
    ~&  & Top-1 & Top-2 & Top-3 \\
    
    \midrule

    \multirow{5}{*}{$w/o$ Text Input} & FaceFormer & 0.236  & 0.351  & 0.413  & 747.6  & 35.03  & 44.20  & 6.973  & \underline{1.745} \\
    & CodeTalker  & 0.238  & 0.351  & 0.422  & 827.9  & 35.04  & 41.21  & 10.48  & 2.438 \\
    & SelfTalk  & 0.293  & 0.407  & 0.482  & 720.1  & \textbf{31.48}  & 43.85  & \underline{6.929}  & 1.775 \\
    & Imitator  & 0.127  & 0.213  & 0.278  & 1315  & 40.24  & 36.60  & 7.474  & 1.920 \\
    & FaceDiffuser  & 0.370  & 0.464  & 0.500  & 221.8 & \underline{32.70}  & 49.74  & 8.466  & 2.046 \\
    & Ours  & 0.404  & 0.520  & 0.584  & 50.51  & 34.66  & \textbf{50.37} & 7.053  & 1.804 \\

    \midrule

    \multirow{1}{*}{$w/$ Text Input}
    & \textbf{Our MM2Face}  & \textbf{0.718} & \textbf{0.854} & \textbf{0.909} & \textbf{41.19} &  34.57 &  \underline{50.09} & \textbf{6.736} & \textbf{1.692} \\

    \bottomrule
    \end{tabular} 
    }
    \caption{Ablation study of text input on benchmark I: text-induced 3D talking head animation.}
    \label{table:bench1}   
\end{table}

\subsection{Text Controlled 3D Talking Head Animation}

\begin{figure*}[h]
\centering
\includegraphics[scale=0.30]{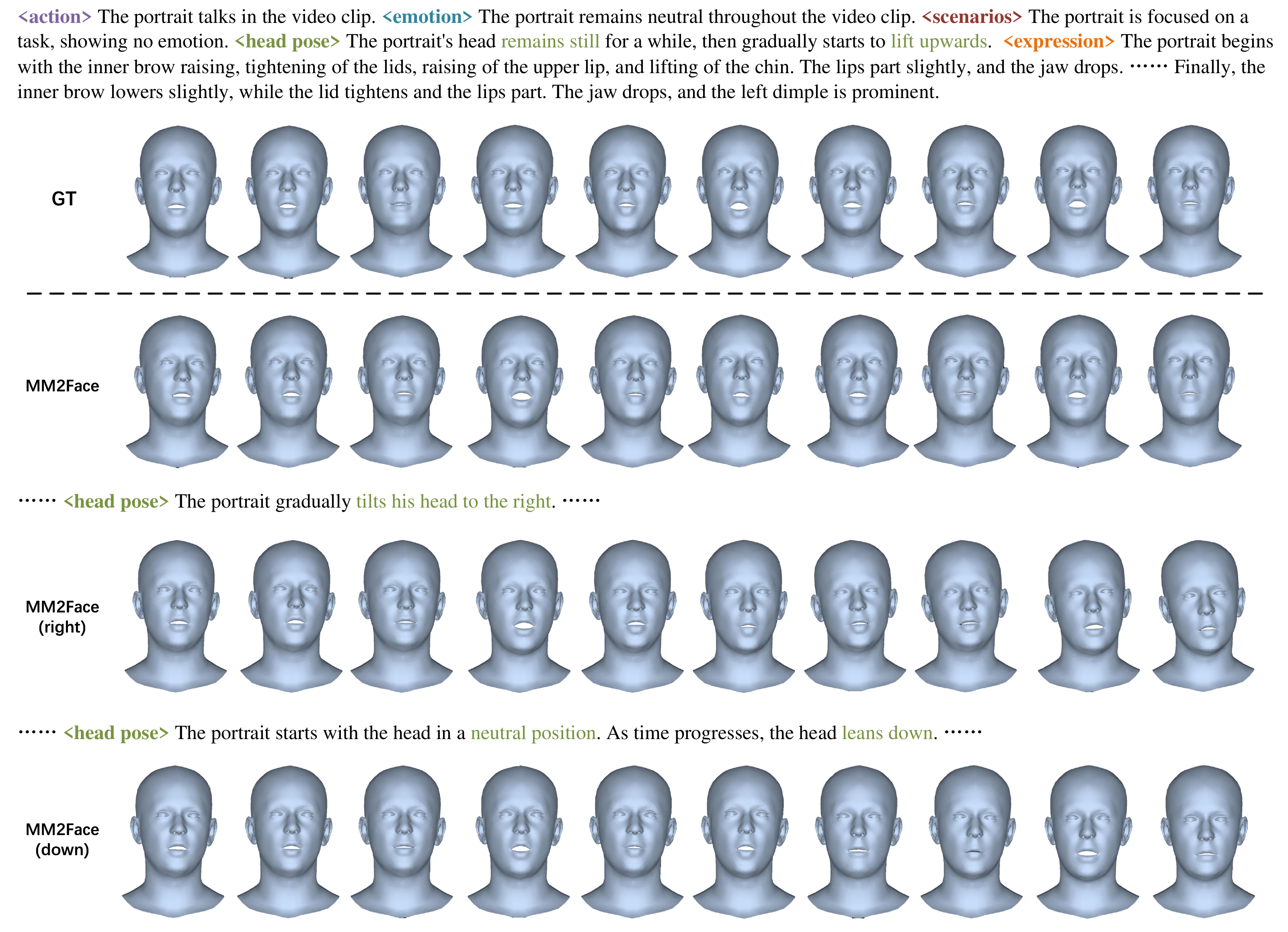}
\caption{Visualization Results of Text guided Head Pose Control of 3D Talking Head.}
\label{fig:headpose_control}
\end{figure*}

We specifically evaluate the influence of text descriptions on generated 3D talking head.

\begin{figure*}
\centering
\includegraphics[scale=0.25]{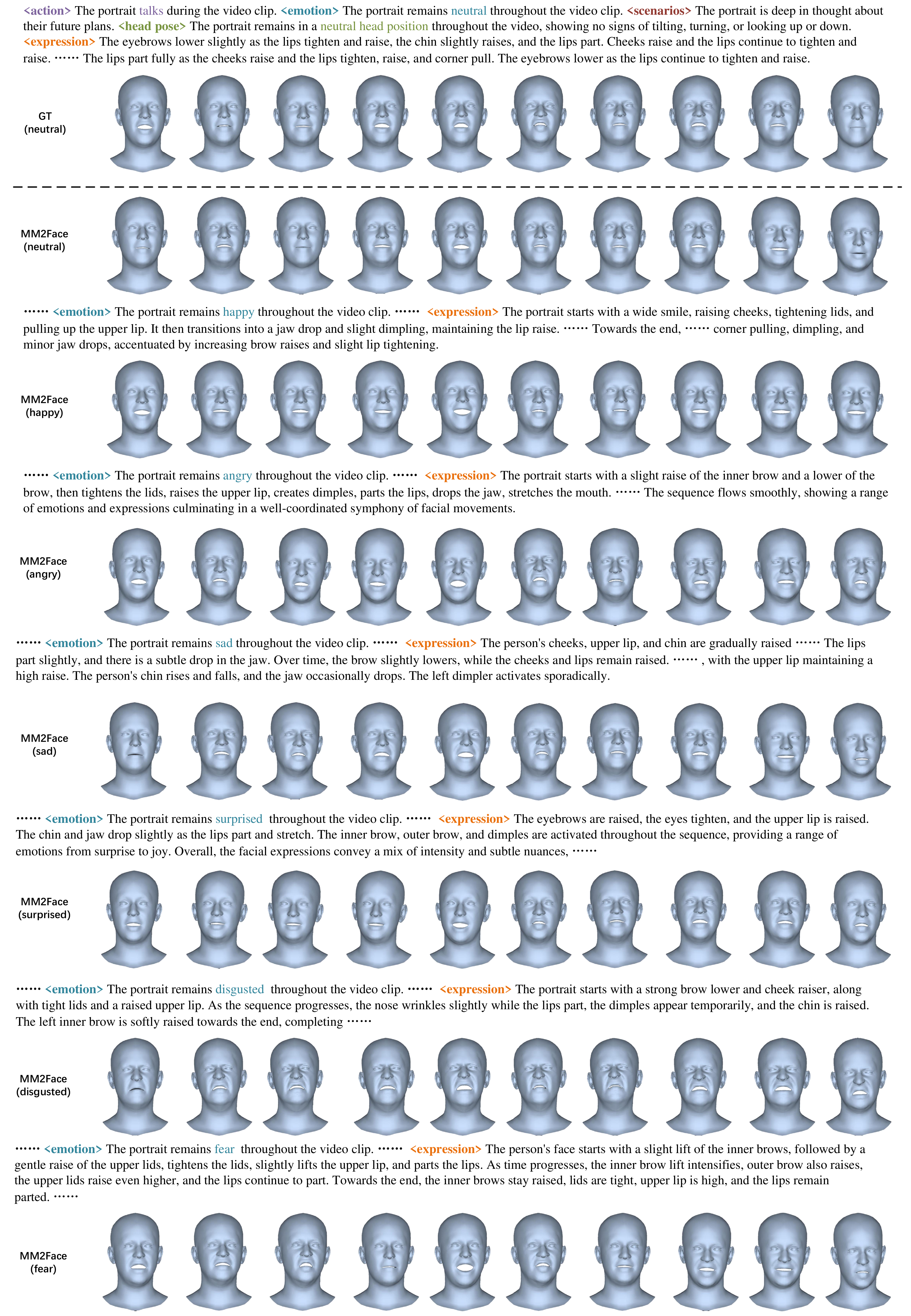}
\caption{Visualization Results of Text guided Emotion Control of 3D Talking Head.}
\label{fig:emo_control}
\end{figure*}

\noindent\textbf{Text guided Emotion Control Results.} We conduct experiments to evaluate text control ability on emotion control. We demonstrate the brilliant emotion control ability of our MM2Face in Fig. \ref{fig:emo_control}. From Fig. \ref{fig:emo_control}, given an audio of a 3D talking head with neutral faces, we show different generation results by changing the abstract emotion and detailed expression descriptions. We select various descriptions from 7 test sequences belonging to 7 different emotions and our MM2Face merges the excellent emotion control ability.

\noindent\textbf{Text guided Head Pose Control Results.} We also provides the text guided head pose results in Fig. \ref{fig:headpose_control}. From Fig. \ref{fig:headpose_control}, we can observe that through changing head pose text descriptions, we can obtain head pose control results given the same audio input.

\section{Limitations and Broader Impacts}
\label{sec:limit}

\noindent\textbf{Limitations:} While our method contributes a satisfactory frame work for both text-induced talking head animation and text-to-3D facial motion generation, our method can still fail to generate some results. First, the reason is probably that for talking head animation, MM2Face model may not disentangle the text conditions and audio conditions perfectly, causing the model sometimes may focus on global text consistency instead of audio consistency. Discovering a more efficient ways to disentangle these two useful control signals is a promising future direction. Second, our method can only animate the FLAME mesh, it is also interesting to animate a high quality head mesh with detailed hairs and faces.

\noindent\textbf{Boarder Social Impact:} There can be also negative impact from this work. Our MMHead dataset and MM2Face model can be used to generate visually realistic and plausible 3D facial motions. And the 3D motions can be further utilized to animate a vivid human avatar. It may be misused to create fake face videos using Deepfake-like technology, which is horrible for society, we sincerely encourage more researchers focusing on deep fake detection.

\end{document}